\documentclass{article}
\usepackage{amsthm}
\usepackage{amssymb}	
\pdfoutput=1	
\usepackage[utf8]{inputenc}
\usepackage{graphicx} 
\usepackage{listings}
\usepackage{wrapfig}
\usepackage{subfigure}
\usepackage{listings}
\usepackage{enumitem}
\usepackage{times}
\usepackage{todonotes}
\usepackage[numbers,sort]{natbib}
\usepackage{soul}
\usepackage{url}
\usepackage{amsmath}
\usepackage{multirow}
\usepackage{url}
\usepackage{float}
\usepackage{caption}
\usepackage{setspace}
\usepackage{floatflt}
\usepackage{times}
\usepackage{graphicx}
\usepackage{listings}
\usepackage{wrapfig}
\usepackage{subfigure}
\usepackage{listings}
\usepackage{enumitem}
\usepackage{times}
\usepackage{todonotes}
\usepackage{soul}
\usepackage{url}
\usepackage{amsmath}
\usepackage{multirow}
\usepackage{url}
\usepackage{float}
\usepackage{caption}
\usepackage{adjustbox,lipsum}
\makeatother
\usepackage{setspace}
\usepackage{floatflt}
\usepackage{times}
\usepackage{authblk}

\usepackage{mwe,tikz}
\usepackage[percent]{overpic}
\usepackage{tikz}
\usepackage{tkz-tab}
\usepackage[pdfborder={0 0 0}]{hyperref}
\providecommand{\keywords}[1]{\textbf{\textit{Index terms---}} #1}

\begin{document}

\title{Semantic Enrichment of Mobile Phone Data Records Using Background Knowledge}

\author[1,2,3,4]{Zolzaya Dashdorj}
\author[2]{Stanislav Sobolevsky}
\author[3]{Luciano Serafini}
\author[4]{Fabrizio Antonelli}
\author[2]{Carlo Ratti}

%

\affil[1]{University of Trento, Italy Via Sommarive, 9 Povo, TN, Italy}
\affil[ ]{\url{dashdorj@disi.unitn.it}}
\affil[2]{Massachusetts Institute of Technology, MIT 77 Massachusetts Avenue Cambridge, MA, USA}
\affil[ ]{\url{stanly,ratti@mit.edu}}
\affil[3]{Fondazione Bruno Kessler, Via Sommarive 18 Povo, TN, Italy}
\affil[ ]{\url{serafini@fbk.eu}}
\affil[4]{SKIL LAB – Telecom Italia, Italy Via Sommarive 18 Povo, TN, Italy}
\affil[ ]{\url{fabrizio.antonelli@telecomitalia.it}}

\date{}
\maketitle

\begin{abstract}

Every day, billions of mobile network events (commonly defined as Call Detailed Records, or CDRs) are generated by cellular phone operator companies.
Latent in this data are inspiring insights about human actions and behaviors, the discovery of which is important because context-aware applications and services hold the key to user-driven, intelligent services, which can enhance our everyday lives.
This potential has motivated preliminary research activities in a variety of domains such as social and economic development, urban planning, and health prevention. 
The major challenge in this area is that interpreting such a big stream of data requires a deep understanding of mobile network events' context through available background knowledge.
Two of the most important factors in the events' context are location and time.
This article addresses the issues in context awareness given heterogeneous and uncertain data of mobile network events missing reliable information on the context of this activity. The contribution of this research is a model from a combination of logical and statistical reasoning standpoints for enabling human activity inference in qualitative terms from open geographical data that aimed at improving the quality of human behaviors recognition tasks from CDRs. We use open geographical data, Openstreetmap (OSM), as a proxy for predicting the content of human activity in the area. The user study performed in Trento city, Italy shows that predicted human activities (high level) from OSM match the survey data with around 93\% overall accuracy, with fuzzy temporal constraints. The extensive analysis of the model validation for predicting a more specific---economical---type of human activity performed in Barcelona city, Spain, by employing credit card transaction data. This data gives us ground-truth information on what types of economic activity are occurring. The analysis identifies that appropriately normalized data on points of interest (POI) is a good proxy for predicting human economical activities, with 84\% accuracy on average. So the model is proven to be efficient for predicting the context of human activity, when its total level could be efficiently observed from cell phone data records, missing contextual information however. 

\keywords{mobile phone data records, big geo-spatial data, human activity recognition, human mobility, context aware computing, qualitative methods, human behavior, semantics, ontology, machine learning, knowledge management, linked open data}
\end{abstract}

\section{Introduction}

Increasingly, massive amounts of data are being generated, stored, and disseminated as a result of human activity. For instance, whenever a mobile phone call, monetary transaction, or social media post is made, geo-located data is automatically generated by mobile network provider, bank, or social network provider (e.g., Facebook or Twitter) and attached to the data record generated by the activity. An extensive body of works leverage for studying human dynamics through cell phone data \cite{ratti2006mlu, gonzalez2008uih, amini2014impact, DBLP:journals/corr/GrauwinSMGR14, sobolevsky2013delineating}, social media posts \cite{hawelka2014geo, paldino2015urban}, bank card transactions \cite{sobolevsky2014money,DBLP:journals/corr/SobolevskySGCHAR14} and vehicle GPS traces \cite{kang2013exploring,santi2014quantifying}. Over the last few years, the use of mobile phones as sensors of human behavior has radically increased. Storage and analysis of information from mobile phones can provide useful insights into how people move and behave. From them, it is possible to infer with a certain level of accuracy, the activities humans are performing at every moment they are connected to the mobile network. Indeed, experiments in large-scale social dynamics have been conducted in the areas of public safety and emergency management \cite{Lu17072012,10.1371/journal.pmed.1001083,DBLP:journals/corr/Pastor-EscuredoMTBWCRLRFOFL14}, health and disease management \cite{Kovanen:Wesolowski_Science2012,6113095}, social and economic development \cite{Frias-martinez12onthe,citeulike:7205422,amini2014impact}, transport/infrastructure \cite{Berlingerio_allabroad,kang2013exploring}, urban planning \cite{5928310,5594641,Girardin_quantifyingurban,citeulike:5158387,10.1371/journal.pone.0037027,pei2014new,DBLP:journals/corr/GrauwinSMGR14}
 and international development, poverty \cite{Smith_ubiquitoussensing} and more \cite{Bogomolov:2014:OUC:2663204.2663254}. A large fraction of mobile phone data has been shown to be extremely useful for humanitarian and development applications (Robert Kirkpatrik UN 2013). This geo---and---time information have explicit or implicit geographic location. 
 
A commonly used source of mobile phone data for which studies is comprised of aggregated and anonymized Call Detailed Records (CDRs), provide meta data about phone activity. This allows us to identify the time, the location, the duration, and possibly of the movements of mobile phones owners. Because the mobile network operator knows the
locations of their cell towers, it is possible to use CDRs to approximate the location of users. The spacing of cell
towers, and thus the accuracy in determining caller's locations, varies according to expected traffic and terrain.
Cell towers are typically spaced 2-3 km apart in rural areas and 400-800 m apart in densely populated areas. Commonly the CDRs are available as anonymous, often incomplete or aggregated because of privacy issues. Accessing personal or private data is a very sensitive and critical topic, which arises claims to be aware of privacy issues before the research outcomes are shared. The main research area of this article is the one, which investigates inferring (collective) human behavior starting from such CDR.


Current research \cite{DBLP:journals/corr/abs-1106-0560, Phithakkitnukoon:2010:AMI:1881331.1881336, Calabrese:2010:GTA:2166616.2166619,Furletti:2012:IUP:2346496.2346500} notes that with a pure CDR, it is possible to identify human behaviors, but results suffer from the heterogeneity, uncertainty and complexity of raw datasets and that the lack of qualitative content is included in the data itself that may be used to help to infer human behaviors. In spite of the bad quality of data, some researchers are able to identify a certain level of human behaviors with the help of features that can be indirectly obtained from the raw CDR. These features are observed information about real-world cases. For example, a user's home location is identified by the frequent stay (location) of the user between midnight to early morning \cite{Calabrese:2010:GTA:2166616.2166619}. These features are mostly specified by domain experts. Most of the behavioral datasets, such as CDR or GPS traces, are relatively large raw datasets where user location is not perfect for tracking the dataset. Such data are incomplete, typically lacking in content and the measurement accuracy of the data is low and coarse grained. While this data provides extensive information regarding when and where people go, it typically does not allow understanding of which activity is performed at that location. 

When the context information of raw data is not directly available in the data, some useful characteristic values for human behaviors can still be discovered by analyzing the activity patterns of mobile network communication \cite{candia-2007,gonzalez08,song2010limits}. For example, anomalous events can be observed by the change of daily communication activity patterns \cite{candia-2007}. Using some semantics about ongoing events (e.g., the location and the time of an event), researchers~\cite{Calabrese:2010:GTA:2166616.2166619,hoteit2014estimating,Furletti:2012:IUP:2346496.2346500,DBLP:journals/corr/abs-1106-0560,Calabrese:2010:GTA:2166616.2166619} are able to discover some relationships between human behaviors (i.e., communication and mobility patterns), and emergency (e.g., earthquake, blackout) or non emergency events (e.g., concert, festival). However, the results are limited to only specific events and the correlations between the events and human behaviors. For instance, the quantitative analysis of behavioral changes in presence of extreme emergency events show radical increase of call frequency right after the event occurs, with long term impacts \cite{DBLP:journals/corr/abs-1106-0560}.

Human behavior is influenced by many external factors, such as weather condition, urban structure, social event. Knowing the context, where a person is important to guess what he/she is doing. For instance, if a person is in or close to a restaurant and makes a phone call around 1:00 pm., or 8:00 pm., then the person is probably looking for eating some food, or (s)he works at the restaurant. In this article, we propose to formalize the context as a pair $\left<l,t\right>$, where $l$ is a location (= geographical area) and $t$ is a time interval. A context can be associated with various information. For instance, weather, events, surrounding geo-objects, and more. For every record in a CDR dataset, it is possible to determine to a certain level of approximation the context associated to such a record. Some research studies~\cite{10.1371/journal.pone.0112608,
10.1371/journal.pone.0045745,10.1371/journal.pone.0045745,10.1371/journal.pone.0081153,
Girardin_quantifyingurban,Phithakkitnukoon:2010:AMI:1881331.1881336} have been performed to understand the correlations of human behaviors to environmental factors, using some additional contextual information about weather, social events and geographical information systems. This focuses on the inference of a certain level of human activities in a certain condition. For instance, human behavior changes during periods of uncomfortable weather condition~\cite{10.1371/journal.pone.0112608}.

Among the literature of raw CDR analysis, researchers mostly employ data-mining and machine learning algorithms\footnote{http://en.wikipedia.org/wiki/Supervised\_learning} \cite{Daniele2009,s141018131} in the analysis. Supervised learning algorithms (e.g., Multiclass SVM, Logistic Regression, Multilayer Perceptron, Decision Tree, KNN) can achieve reasonable accuracy on the CDR analysis but they require qualitative and quantitative properties in data and ground truth information in order to perform behavior classification well. The researchers claimed to acquire a sufficient amount of labeled training data. For their studies, but obtaining ground-truth information (e.g., user diaries and surveys) for training data remains a very expensive and almost unmanageable task, especially when one considers the large amounts of CDR data that need to be annotated and the diversity in the contexts of mobile phone events that makes the annotation more difficult. 

The result of CDR analysis usually provides quantitative evidence that are presented in visual analytic tools (e.g. graphs and tendencies), and requires a qualitative description for the human behavior interpretation. Less interest has been dedicated to the development of methods that are capable of producing a qualitative/semantic level description of human behavior. This thesis will concentrate on providing qualitative descriptions of human behavior. For representing human behavior, we propose to use semantic descriptions based on an ontology. More in details, a semantic description of human behaviors is a representation of the behaviors of a single person, the behaviors of a group of people or the events that happen in the human society, in terms of concepts and relations of an ontology. An example of a semantic description of human behaviors is in the fact that ``a person is performing some specific actions'' (e.g. working, shopping and hiking), or the fact that ``certain events are happening in a certain area'' (e.g. a car accident and a train suddenly stopping in the middle of nowhere).


Growing information in a variety of Web 2.0 and social platforms opens opportunities for collecting and using information about the context associated to mobile phone data records. These sources include a wide and diverse set of data that can be useful to characterize a user's context: environmental data; statistic description of the territory; public and private, emergency and non-emergency events; statistical data about demography, ethnography, energy or water consumption; and more. These kinds of data should be employed as contextual information if we are to gain a better understanding of a user's context, as they allow a more representative and semantically expressive characterization of the relationships between human behavior and different contextual factors. 

This article seeks understanding human behavior based on contextual information obtained (also) by leveraging available Web 2.0 data sources, representing them in both a conceptual and computational model, such as the correlations between contexts and human behaviors, and prediction of human activities of a group of people will perform in a given context, and the most probable action that a group of people are doing in a context. The contextual information can be inaccurate, unavailable, uncertain and noisy. In our research, we need to pre-estimate if the information is useful enough by measuring the validity, accuracy, and suitability of the data. Such contextual data needs to be effectively integrated, coping with the problems caused by uncertainty and heterogeneity of data.

We identified that points of interest (POIs) are good proxies for predicting the content of human activities in the context in which mobile network events occur. Our proposed model is named High Level Representation of Behavior Model (HRBModel), and correlates POIs and the time of the day with typical user activities. The model integrates an ontology for POIs and times/days with an ontology for human activities and statistical information about the correlations between the two. Given a set of POIs and a time of the day (which can both be inferred from the context), the model generates a set of human activities associated with a likelihood measure.

We validated the accuracy of the model using two different qualitative geo-located data-sources, providing us ground-truth information on what type of human activities were performed in a given location: 1) user feedback collected in Trento, Italy and, 2) bank card transaction data generated in Barcelona, Spain. Our extended evaluation of the model includes a validation of the impact of heterogeneity and uncertainty of open geographical data for recognizing human activities. This validation is performed on city scale and takes different land-use types into account in order to understand the level of accuracy of human activity prediction from the geographical data or vice versa from bank card transaction data. 

The innovative aspect of the research is the development and usage of ontologies for analysis of raw CDR data, combined in a mixed model with data mining numerical methods. Our method improves the quality of human activity recognition tasks given noisy, lossy, and uncertain data, and allows to infer, with a certain level of accuracy, the different level (of hierarchy) of activities humans are performing at a specific location and time. Accordingly, the model can deal with a wide range of contextual features about objects of a territory by extracting concepts and inter-concept relations. 

Another innovative aspect is that the HRBModel is a general model that can be used to provide semantic interpretation of any type of geo-located and time-dependent data, other than CDR, such as data provided by social platforms like Foursquare, or Twitter, GPS, and Credit Card data. Potential applications of this research are a context aware application systems, and  ``smart cities'' applications that provide decision support for stakeholders in areas such as urban, transport planning, tourism and event analysis, emergency response, health improvement, community understanding, economic indicators and others ~\cite{nrc12377,DBLP:journals/corr/abs-1210-0137,DBLP:journals/pervasive/FerrariBCC13}. 

The remainder of the paper is structured as follows: Section \ref{related works} briefly summarizes relevant literature. In the following section, we introduce the data-sources we used for contextual information. The core methodology of the HRBModel to extract high level human activities from geographical information is explained in Section \ref{methodology}. The experimental results and evaluation of the model are presented in Section \ref{evaluation}. In Section \ref{usage of the model}, we discuss the usage of the model and describe possible developments (extensions). Finally, in Section \ref{conclusion}, we draw some conclusions.

\section{Related Works}\label{related works}

Great opportunity and impact on human activity recognition is brought by mobile phone data. Through data mining and machine learning techniques, human activity data becomes available for further analysis. 

Researchers in the areas of behavioral and social sciences are interested in examining CDR to characterize and understand real-life phenomena, such as individual traits, human mobility \cite{DBLP:journals/corr/abs-1106-0560,Calabrese:2010:GTA:2166616.2166619,4756329,Timothy-2006}, communication and interaction pattern\cite{Calabrese:2010:GTA:2166616.2166619,
onnela-2006,10.1371/journal.pone.0014248}. Candia et al~\cite{candia-2007} proposes an approach to understand dynamics of the individual calling activities, which could carry implications on social networks. The author analyzed calling activities of different group of users; (some people rarely use mobile phone, others use often). The cumulative distribution of consecutive calls made by each user is measured within each group and the result explains that the subsequent time of consecutive calls is useful to discover some characteristic values for the behaviors. For example, peaks occur near noon and late evening. The fraction of active traveling population and average distance of travel are almost stable during the day. This approach can be applied for detecting anomalous events. In \cite{Calabrese:2010:GTA:2166616.2166619}, the authors analyze the mobility traces of groups of users with the objective of extracting standard mobility patterns for people during special events. In particular, this work presents an analysis of anonymized traces from the Boston metropolitan area during a number of selected events that happened in the city. They finally proved that people who live close to an event are preferentially interested in events organized in the proximity of their residence. 

Some researchers used semantic tags for geographic location from social networks or user diaries to identify a semantic meaning of places \cite{Ye:2011:SAP:2020408.2020491,conf/huc/KrummR13,
Sakaki10earthquakeshakes,Sengstock:2011:ECG:2093973.2094017,
Yin:2011:GTD:1963405.1963443} captured in user CDRs or GPS trajectories. Supervised learning algorithms (e.g. binary SVM and hidden markov models) are employed for this identification.

People perform different activities even when they stay in the same location at the same time, with respect to the wide difference of situations caused by various factors like natural, technological or societal disasters, from hurricanes to violent conflicts. This requires the consideration of potential influencing context factors on a common space-time basis that enables the associations of those different datasets. Indeed, some researchers~\cite{DBLP:journals/corr/abs-1106-0560, Phithakkitnukoon:2010:AMI:1881331.1881336, Calabrese:2010:GTA:2166616.2166619,Furletti:2012:IUP:2346496.2346500} use an additional information about context factors in order to study the relationship between human behaviors and context factors like social events, geographical location, weather condition, etc. This is always as successful as the quality of the context factors.

In fact, some researchers validated the use of spatial and temporal contextual features for understanding the relationship between human behaviors and environmental factors. For instance, Sagl and Resch et al. \cite{Sagl2014} found that many factors influence the collective human behavior and weather conditions and geographic information are certainly two of these factors. Therefore, associating environmental and social factors to mobile phone data is extremely useful for analyzing the dependency of human behaviors from the external factors. These studies focus on the inference of a certain level of human activities under certain conditions like human behavior changes in uncomfortable weather conditions~\cite{10.1371/journal.pone.0112608,10.1371/journal.pone.0045745,10.1371/journal.pone.0045745,10.1371/journal.pone.0081153}, 
human mobility of different communities~\cite{Phithakkitnukoon:2010:AMI:1881331.1881336,candia-2007}, human mobility during an  event~\cite{Calabrese:2010:GTA:2166616.2166619,D4D:2013,
DBLP:journals/corr/abs-1106-0560}, or communication activity patterns in different land-use types~\cite{Girardin_quantifyingurban,Phithakkitnukoon:2010:AMI:1881331.1881336,
pei2014new,sobolevsky2013delineating}. In this work, geographical information in particular points of interest (POIs) are collected using pYsearch (Python APIs for Y! search services) from a map. They annotated POIs with four type of activities; eating, recreational, shopping and entertainment. The authors analyzed human activity patterns (i.e., sequence of area profiles visited by users) correlating to geographical profiles. Bayesian method is used to classify the areas into crisp distribution map of activities, which enables the activity pattern extraction of the users. The results shows that the users who share the same work profile follow the similar daily activity patterns. But not only these area profiles can explain the mobility of these users and the enlargement of activity or event taxonomy in those areas can enable the classificaions of these activity patterns. Due to the limitation of heterogeneity of activities considered in this research, the results in measurement lacks of certainty. 

The authors of \cite{Calabrese:2010:GTA:2166616.2166619} find that residents are more attracted to events that are organized close to the home location of the residents. F. Girardin et al.~\cite{Girardin_quantifyingurban} analyzed aggregate mobile phone data records in New York City to explore the capacity to quantify the evolution of the attractiveness of urban space and the impact of a public event on the distribution of visitors and on the evolution of the attractiveness of the points of interest in proximity. In \cite{DBLP:conf/icsdm/SaglBRB11}, the authors introduced an approach to provide additional insights in some interactions between people and weather. Weather can be seen as a higher-level phenomenon, a conglomerate that comprises several meteorological variables including air temperature, rainfall, air pressure, relative humidity, solar radiation, wind direction and speed, etc. The approach has been significantly extended to a more advanced context-aware analysis in \cite{s120709800}.

\section{Data gathering}\label{data-source}

In this study, we model one contextual location dataset of 4.6 million POIs in Trento, Italy, 2.6 million POIs from OSM in Barcelona, Spain against two 130K POSs from Banco Bilbao Vizcaya Argentaria (BBVA), and CDRs from GSM 900 and GSM 1800 (Telecom Italia). Here, we describe the structure of mobile network events and to investigate how mobile network data can be combined with the available contextual information for supporting human recognition task. We show that location and time are the key factors to integrate mobile phone data records with the contextual information coming from the other sources about environments and events. 

\subsection{Mobile phone data records}

There are many possible types of CDRs generated by operator companies. In the followings we describe the ones we are using to use
in this article. 
The initial experiments of this thesis have been done on a CDR dataset (completely anonymized)
provided by Telecom Italia with the following structure:
\begin{description} 
\item[source cell and target cell:] the identifiers of the cell from
  which the call is issued and the receiver's cell, respectively. They are
  composed of a Location Area Code and additional parameters, such as
  the operator, etc. Cell ID needs to be resolved into a physical
  geographical area
\item[date and time]
The date and the time of the day in which the call is issued
\item[duration]
The duration of the call
\end{description}

To combine mobile phone data records with contextual information, location and time associated to each single call are key. CDR are not directly geo-referenced but they are labelled with the cell. From the cell, and the geographical cell distribution we can reconstruct with a certain level of accuracy and estimation of the geographical location of the user when he or she generated a particular mobile event.

As an example, here we describe the structure of cell coverage map that generated by Telecom Italia. We use aggregate GSM 900 and 1800 data that was provided by Telecom Italia. The cell coverage map of the mobile network is captured across Trento containing completely anonymized cell (antennas). Cell identification is used for identification of some portion of a physical geographical area featured within a set of devices (antennas) that support the mobile communication (e.g. calls, sms and internet). The size of coverage areas can vary depending on the estimation of
call traffic from/to this area. Usually the size of the coverage areas
are inversely proportional to the density of the population inhabiting
the area. However
there is an upper-bound of the coverage area size due to antennas
physical limits to 35 km. The cell partition is not fixed and can vary
depending on the demand of network during specific periods. For
instance to support an increased request of communications, additional
cells can be dynamically activated in order to increase the capability
of the entire network. Partial knowledge of mapping between cell ID's
and coverage areas, irregularity of the size of the cell, and
dynamically of the cell configuration add complexity and uncertainty
on the identification of the physical location from where a call is
issued. Nowadays, cell coverage maps are restricted to be accessed due
to privacy issues. Yet, some open sources for cell coverage maps are
available on the web sites, such as Open
Cell-ID\footnote{http://www.opencellid.org} and Open Signal
Maps\footnote{http://opensignal.com}

\subsection{Credit card data}

Credit card data is raw data semantically enriched with some semantic information which describes the type of transaction that will be used in our analysis in order to evaluate the proposed approach. The descriptions of which are provided for research purposes by one of the largest Spanish banks, BBVA. The raw dataset is protected by the appropriate non-disclosure agreement and is not publicly available. The data was completely anonymized on the bank's side, preventing access to any personal information of the customers in accordance with all applicable privacy protection regulations. Each transaction is characterized with its value, a timestamp, and additionally the retail location, where it was performed at the Points of Sale (POS) terminals. Each POS is categorized into one of the 76 business activity types, such as groceries, fashion, bars, restaurants, sport shops, etc.

\subsection{Contextual data}

Openstreetmap (OSM) is an open and free map of the world enriched with a large number of geo-referenced objects upon an open content license. The database of OSM is frequently updated and is rapidly growing. It has a free tagging system based on a well documented taxonomy, which classifies objects in categories, such as roads, buildings, etc. OSM adopts a topological data structure based on the three core elements:

\begin{description} 
\item[Nodes:] points with a geographic coordinate (i.e., a pair of latitude and longitude). A node mostly describes a POI. For example, shops, restaurants, etc. 
\item[Ways:] ordered lists of nodes in a poly-line or a polygon. These elements are used for representing linear features or areas, such as streets and forests. 
\item [Relations:] groups of nodes and ways that represent the relations among existing nodes and ways, such as roads with several exiting ways. 
\end{description}

Each of aforementioned three element types is described with geographic coordinates. Nodes are mainly used to represent POIs, but they can also be used as part of ways to represent linear features. So a POI can be described in a node or in nodes, and part of a way, for instance, a railway level crossing. The metadata of OSM is provided in the form of tags (i.e., a textual description described as a pair of key and value) to describe each element. Nodes, ways, and relations are tagged, and that the tag represent some information about the node, for instance if it is a station, a shop, a road, a crossing, etc. A list of common tags used (but not all) is captured in the Map Features page\footnote{http://wiki.openstreetmap.org/wiki/$Map\_Features$}. The tags for nodes, ways, and relations are represented as a set of pairs of $\left<key,value\right>$. For example, a university is tagged as that the key is amenity, the value is university, and the highway route is tagged as that the key is route, the value is highway. The OSM tags for one node corresponding to a supermarket in XML representation are shown below:: 
\begin{verbatim}
<node id="618033185" version="3" uid="330007" 
  user="pikappa79" 
  timestamp="2011-05-26T16:02:14Z" 
  changeset="8254868" 
  lat="46.0946011" lon="11.1162507">
    <tag k="addr:postcode" v="38121"/>
    <tag k="shop" v="supermarket"/>
    <tag k="addr:country" v="IT"/>
    <tag k="name" v="Supermercato PAM"/>
    <tag k="addr:housename" v="Bren Center"/>
    <tag k="addr:street" 
    v="Via Giovanni Battista Trener"/>
    <tag k="addr:city" v="Trento"/>
</node> 
\end{verbatim}

The way we intend to use OSM in this article is by collecting the objects which are in the cell area. These objects can provide insights on the type of action a person while he is located in a cell area. For instance, if a person is in or close to a restaurant and makes a phone call around 1:00 pm., or 8:00 pm., then the person is probably looking for eating some food. Alternatively or complementary, we could use Google Map or Yahoo Map or regional cadastral map database for collecting the POIs within the each cell area. A relevant technique for POIs extraction in a given region from geographical information system is described in~\cite{DBLP:conf/mum/DashdorjSAL13,Phithakkitnukoon:2010:AMI:1881331.1881336}.

Table \ref{table:ontology mapping} introduces a brief description of ontology we used to generate useful information from the aforementioned heterogeneous data sources. For the integration between these qualitative data-sources we designed an ontology for each domains used in this PhD research work: 1) POI, 2) Action, 3) Time. This ontology is based on surveys, crowdsourcing, and domain experts, or the formally defined classifications for instance, business career classification by YellowPages or sport classification by the Olympic Game Committee, etc. 

\begin{table*}[!htbp]
\centering
    \caption{Input and Output cardinalities}
	\label{table:ontology mapping}
	\begin{adjustbox}{max width=\textwidth}
    \begin{tabular}{|l|l|l|l|l|}
        \hline
        \textbf{Ontology} & \textbf{Input}  & \textbf{Input}  & \textbf{Output}  & \textbf{Output} \\ 
        &  & \textbf{Cardinality} &  & \textbf{Cardinality} \\ \hline
        OSMonto & OSM & 1822 & POIs & 517\\ \hline
        ActOnto & Survey, Crowdsourcing & 420 & actions & 217\\ 
            & YellowPages  &  &  &  \\ \hline
        TimeOnto & Survey, experts & 72 & fuzzy times & 27 \\ 
                    &   &  & days &  \\ \hline
        HBOnto & OSMonto, ActOnto & 52 classifiers & actions & parent-10 \\ 
                & TimeOnto &  &  & all-217 \\\hline

    \end{tabular}
    \end{adjustbox}
\end{table*}

\section{High-level representation of behavior model}\label{methodology}

This section introduces a prediction model, called High Level Representation of Behavior Model (HRBModel)~\cite{DBLP:conf/mum/DashdorjSAL13}. This model generates a set of human activities with a likelihood from the set of POIs. The structure of the model is visualized in Figure \ref{gra:framework}. The HRBModel consists of two core components: the Human Behavior Ontology (HBOnto) and the Stochastic Behavior Model (SBM). The HBOnto is an ontology that provides a classification of POIs, human actions, and periods of the day, and it formalizes the relations between these three sets of elements. For example, at the university, people are studying in the morning and afternoon. The SBM provides a likelihood measure to each of the actions that has been selected by the ontology, on the basis of the frequency of POIs and on a fuzzy model of the actions depending on the time of the day. It allows us to predict the possible top-$k$ activities associated with a likelihood, that could be performed in each context of a given region. For example, having a breakfast activity is the most probable activity at residential types of area in the early morning.

In the remainder of the chapter, we describe in details the two components of the HRBModel: HBOnto in Section \ref{HBONTO} and SBO in Section \ref{SBO}. In Section \ref{sec:application HRBModel}, we show the application of the HRBModel.

%
%
%

\subsection{Human behavior ontology (HBOnto)}\label{HBONTO}

\begin{figure*}[htb!]
  \centering
  \subfigure[The prototype framework of the HRBModel]
  {\includegraphics[scale=0.22]{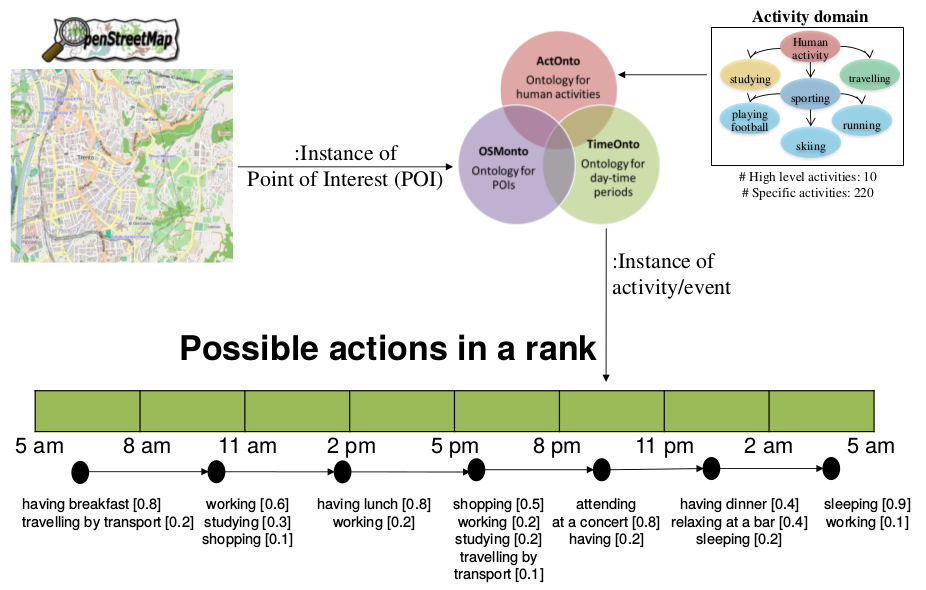}
  \label{gra:framework}}
  \quad
  \subfigure[Composition of human behavioral ontology (HBOnto)]
  {\includegraphics[scale=0.26]{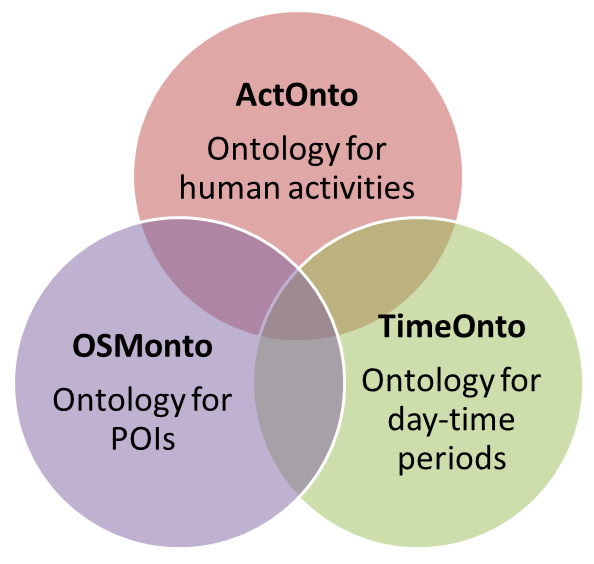}
  \label{gra:HBOnto}}
  \caption{}
\end{figure*}

The Human Behavior Ontology (HBOnto) provides a set of possible activities for a given POI type or in the time of day, i.e., it does a priori activity selection on the basis of POIs and time of the day. For instance, it excludes ``sleeping'' if the time of the day is 1:00 pm. and it excludes ``hiking'' if you are in the city center. It includes activities only if there are POIs that activates them. For example, an area containing a swimming pool activities swimming activity. The HBOnto\footnote{http://brenta.disi.unitn.it/~dashdorj/Ontology/HBOnto.zip} is composed of three ontologies (see Figure \ref{gra:HBOnto}). OSMOnto, which contains the various classes of POIs as classified in OSM; ActOnto, which contains the different types of human activities we are interested in; TimeOnto, which contains temporal references in which those human activities can occur. These three ontologies are related by two main relations. The first one connects an action class to the classes of POIs, in which the action can be performed. The second relation connects an action class to the classes of the time of day during which the action is possible. In the following, we describe each ontology of HBOnto (Sections \ref{sec:OSMONTO}, \ref{sec:ActOnto}, \ref{sec:TimeOnto}), the relations among them (Sections \ref{sec:relation POI and activity} and \ref{sec:relation time/day and activity}) and how DL reasoning on HBOnto can be used to determine possible activities at a given POI type and time (Section \ref{sec:reasoning}).


\subsubsection{The Ontology of POI---OSMonto} \label{sec:OSMONTO}

\begin{figure*}[htb!]
       \centering
       \includegraphics[scale=0.35]{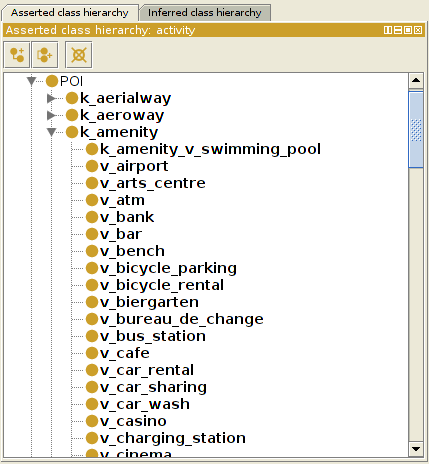}
       \caption{OSMOnto ontology for POI classes}
       \label{gra:osmonto classes} 
\end{figure*}

A POI denotes a geo-referenced object, such as a restaurant, a shop,
or a lake or other objects, whih have a precise geographical location. In this study, we extended the existing OSMonto ontology~\cite{CodescuOSMONTO:2011:DAS:2008664.2008673}, which proposes a classification of POIs, by adding new POI types that have recently been appended to OSM (see, for instance~\cite{osmwiki}) as well as the new frequent POI types found in our analysis. The (extended) OSMonto ontology is a taxonomy of POI types, that formalizes the information about POIs contained in OSM. The POIs are encoded in the OSM as in the form of a pair $\left<key,value\right>$.
This information is transformed in order to be represented in OSMOnto. The key is a type of POI that includes a prefix ``$k\_$'', and the value is a sub type of POI that includes a prefix ``$v\_$''. However, according to the OSM tagging structure about POIs, some POIs might have the same ``$v\_$'' value denoting different sub-types based on the ``$k\_$'' key. For instance, a POI $\left<railway,train\right>$, and a POI $\left<route,train\right>$ both have the same values. In this case, we represent those types of POIs as the combination of the key and the value, for instance in $k\_railway\_v\_train$. An example of the classes of POIs in OSMOnto is shown in Figure~\ref{gra:osmonto classes}. We added manually the recent POI types appeared in OSM. In total, 517 POI types are used in OSMonto. We discard and exclude the irrelevant POIs, for example, those do not reflect a human action like benches, chimneys, power towers. and trash bins. 

\subsubsection{The Ontology of Human Activity---ActOnto} \label{sec:ActOnto}
\begin{figure*}[htb!]
       \centering
       \includegraphics[scale=0.30]{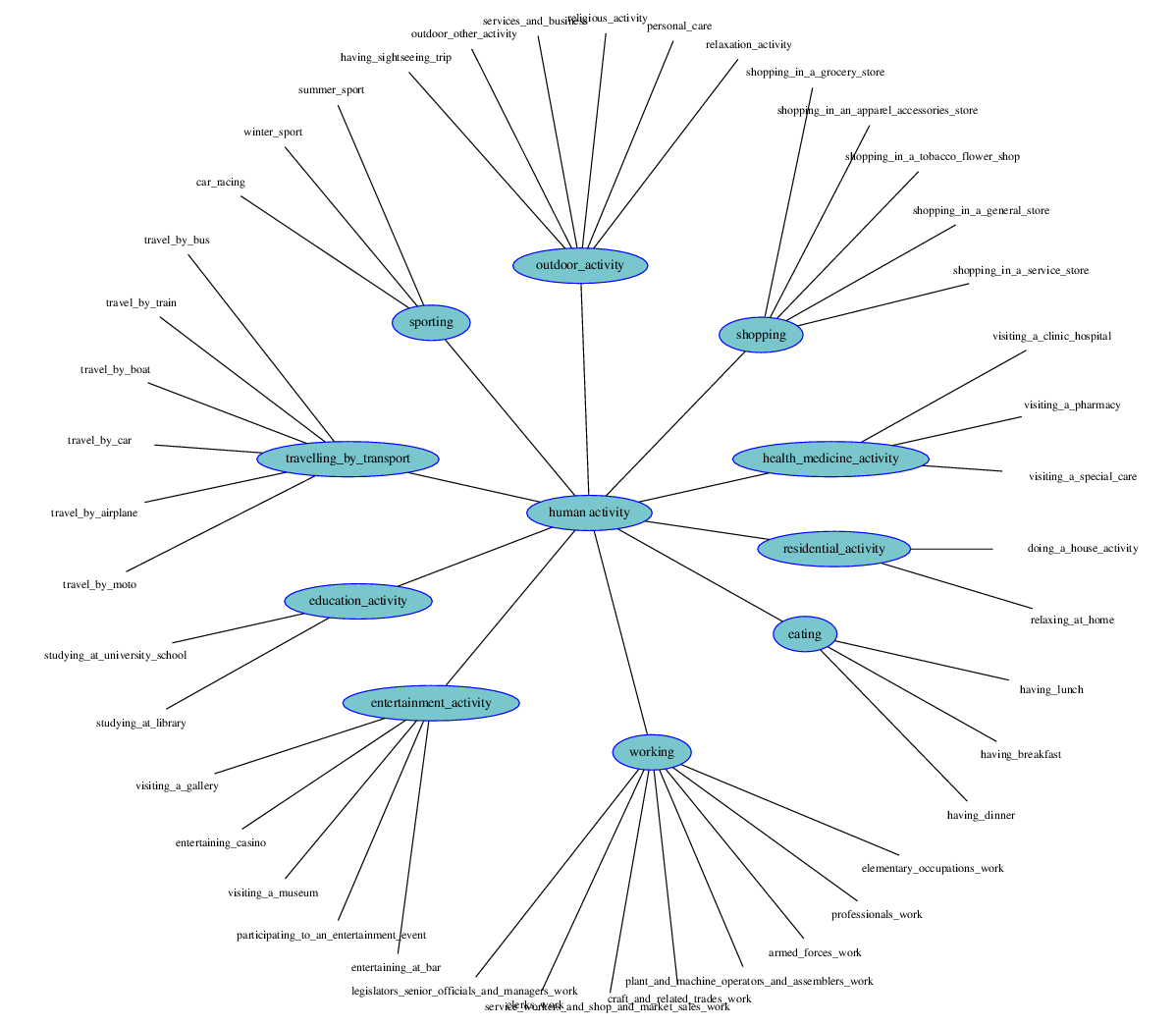}
       \caption{Hierarchy of human activity classes are visualized in two levels down from the top level of human activity classes in ActOnto in which the nodes are the top level of activities and the leaves are the second level of activities}
       \label{gra:activity_hierarchy} 
\end{figure*}

The ActOnto ontology is a taxonomy of human activities. ActOnto contains totally 217 human activity classes hierarchically organized in different levels up to four levels. The upper level is composed of 10 activity types (see the Table~\ref{table:group action}). Various sources are exploited to classify those activities, such as Yellow Pages\footnote{http://www.yellowpages.ca/business/}, Human Resource Management\footnote{http://mayor2.dia.fi.upm.es/oeg-upm/files/hrmontology/hrmontology-RDF.zip}, and Olympic Games\footnote{http://swat.cse.lehigh.edu/resources/onto/olympics.owl} and we also asked for support from domain experts. The ontology has been constructed manually. Figure \ref{gra:activity_hierarchy} shows the top two levels of the human activities taxonomy. We use description logic to formalize the hierarchy of human activities with axioms. For example, $travel\_by\_airplane \sqsubseteq travel\_by\_transport, travel\_by\_car \sqsubseteq travel\_by\_transport\\ $.

\begin{table*}[!htbp]
\centering
    \caption{Human activity categories in ActOnto}
	\label{table:group action}
	\begin{adjustbox}{max width=\textwidth}
    \begin{tabular}{|l|l|}
\hline
        \textbf{Top level classes of activities} & \textbf{Type of POIs} \\ \hline
        eating & fast food, food court, restaurant, cafe\\ \hline
        shopping & grocery, general stores\\ \hline
        health medicine activity & hospital, pharmacy\\ \hline
        entertainment activity & bar, casino, movie, theater\\ \hline
        education activity & library, university school\\ \hline
        transportation traveling & airplane, bus, car, train\\  \hline
        outdoor activity & sightseeing, personal care, religious places\\ \hline
        sporting activity & car racing, summer, winter sports\\ \hline
        working activity & professional work place, industrial place\\ \hline
        residential activity & guest house, hotel, hostel, residential building\\ \hline
    \end{tabular}
    \end{adjustbox}
\end{table*} 

\subsubsection{The Ontology of Time---TimeOnto}\label{sec:TimeOnto}

Human activities are also correlated to time and, in particular, to the time of day and the day of week. For example, in early morning, people have breakfast and in the afternoon people have lunch. For the task of activity detection, we are more interested in a qualitative representation of time than fine-grained time measurements. The TimeOnto ontology is designed for modeling time in qualitative terms (e.g., late morning), organizing them into a containment hierarchy (e.g., morning includes late morning). The qualitative terms fuzzify numerical time periods. A representation of how fuzzy times/days are hierarchically organized is given by following axioms in description Logic:

\begin{itemize}[]
\itemsep0em 
\setlength{\itemsep}{0pt}
    \item[] $morning \sqsubseteq time$
    \item[] $early\_morning \sqsubseteq morning$
    \item[] $mid\_morning \sqsubseteq morning$\\\ldots
    \item[] $weekday \sqsubseteq day$
    \item[] $holiday \sqsubseteq day$
    \item[] $Saturday \sqsubseteq holiday$
    \item[] $Sunday \sqsubseteq holiday$\\\ldots
\end{itemize}

In total, 17 classes of fuzzy time and 10 classes of fuzzy day are described in the ontology.

\subsubsection{Relation between POIs and Human activities} \label{sec:relation POI and activity}
A POI activates a set of human activities that people can perform when they are in that object, or in its neighborhood. For instance, in a restaurant people usually perform eating activity. On a highway road, traveling by transportation is the most usual activity. Every $poi_i$ is related with a set of activities $A(poi_{i})  \sqsubseteq \{a_{1}, a_{2},.., a_{n}\}$ which can be performed or hosted there or nearby. For example, on a highway road, \textit{traveling by car} is the most usual activity. The relationship between human activities and POIs is represented with the object property ``$what\_can\_be\_done$''. The assertion of this relationship~\cite{FOST} is an existential restriction that is exemplified by the following axiom, which states that for every highway road there is an activity of the type \textit{traveling by car}, which is possible: $ v\_highway \sqsubseteq\exists what\_can\_be\_done.traveling\_by\_car $. As a result, we defined around 151 associations between human activities and corresponding POI types. 

\subsubsection{Relation between Human activities and Time periods} \label{sec:relation time/day and activity} 

The activities of a user are highly influenced by his location. For instance, if a person is close to a university, the most probable activities are studying, teaching, and working. In order to capture this dependency, we first need to model which activities can be performed or hosted within or nearby every POI (e.g., eating is possible in a restaurant, while traveling is possible on a railway). Different activities have different timetables; these timetables may depend on a particular context/culture, e.g., the city of Trento. Then we performed the experimental evaluation. 
\begin{itemize}
\itemsep0em 
\item shopping activity [9 :00 am,12:00 pm.], [3:00 pm.,6:00 pm.], except Sunday.
\item eating activity [8:00 am.,2:00 pm.], [6:00 pm.,10:00 pm.].
\item entertaiment activity [8:00 pm.,3:00 am.].
\end{itemize}

Every human activity type is related with a set of times/days $A(t_i)  \sqsubseteq \{a_1, a_2,.., a_n\}$. This relationship between time of the day and human activities is represented with an object property, ``$is\_usually\_done\_during$'' and the relationship between days and human activities is designed as an object property ``$is\_usually\_done\_on$''. The assertion of these relationships is an existential restriction, which is specified using the following example axiom in DL: 
$ shopping \sqsubseteq activity \sqcap \exists is\_usually\_done\_during.(morning \sqcap afternoon)  \sqcap \exists is\_usually\_done\_on.(weekday \sqcap saturday)$. This axiom should be read as ``every shopping activity can be possibly performed in the morning and in the afternoon except on Sunday''. We attempt to associate time periods to all actions using the above type of axiom accordingly. As a result, all human activities in HBOnto are associated with the day of the week and the dime during the day. 

\subsection{Usage of the HBOnto ontology}\label{sec:reasoning}

We explain the formulas of the ontology for deriving of a set of human activities that could be performed in a certain POI during a certain time of the day and day of the week. We use an axiom for deriving the activities. We use a DL reasoner, for instance, FACT++ reasoner in Pellet\footnote{http://clarkparsia.com/pellet/}. 

\subsubsection{Deriving Human Activities in a Given POI type} 

For deriving a set of human activities from a certain POI, there is a boolean validity function VAL$(poi_i, a_i)$ that indicates whether an activity is valid or is not in $poi_{i}$ which exploits the reasoning of the axioms defined, as the following example axiom: $v\_highway \sqsubseteq\exists what\_can\_be\_done.traveling\_by\_car$.

\subsubsection{Deriving Human Activities at a certain time and day} 
For extracting a set of human activities in a certain time of the day and day of the week, there is a validity boolean function VAL$(a_i, t_i, d_i)$ that indicates whether an activity is valid or is not at $t_i$ in $d_i$. The function is derived from the reasoning axioms called derived facts. For example, in order to derive what are the activities that are usually done during the early afternoon of workdays as the following example axiom: $workday\_early\_afternoon\_activity \equiv activity  \sqcap \\ 
\exists is\_usually\_done\_during.early\_afternoon \ \sqcap  \exists is\_usually\_done\_on.workday$

\begin{figure} \centering
\begin{tikzpicture}[      
        every node/.style={anchor=north east,inner sep=0pt},
        x=-2mm, y=-2mm,
      ]   
     \node (fig1) at (0,0)
       {\includegraphics[scale=0.40]{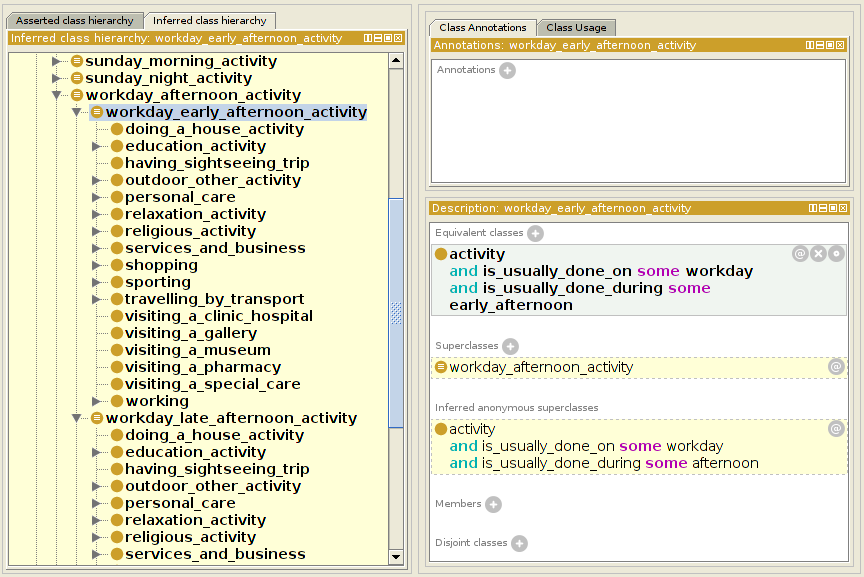}};
     \node (fig2) at (3,3)
       {};  
\end{tikzpicture}
\caption{Example of a classifier atomic formula for human activities on weekday early afternoons}
\label{gra:reasoning}
\end{figure}

We modeled 52 axioms for this reasoning. For example, a reasoning on human activities in the early afternoon of weekday is shown Figure \ref{gra:reasoning}.

\subsection{Stochastic behavior model (SBM)}\label{SBO}

The HBOnto ontology provides a set of possible actions, a stochastic model (SBM) - the other component of the HRBModel we propose - tries to rank the actions on the basis of the number of POIs of certain type and the distance from them, and a fuzzy model from the time of the day. Thus we compute as following: $P(a|l,t,d) = \frac{P(a|l) * P(a|t,d) }{P(a)}$, where conditional independence of $l$ and time periods $\left<t,d\right>$ in given $a$. Section \ref{sec:action in time} explains the estimation of the $P(a|t,d)$ based on the relative importance of action in certain time period $t$ of $d$ employing Fuzzy model. Section \ref{sec:action in location} describes the estimation of the $P(a|l)$ based on the relative importance of POIs in location $l$ employing TF-IDF. Section \ref{sec:nearby location} shows how this approach can be generalized to $P(a|l,r)$ in order to account for POIs that are in a larger neighborhood of the user to make the model more robust. $P(a)$ is the total probability of actions occur in all locations and all certain time periods of all days.

\subsubsection{Likelihood of human activity in a given time/day}\label{sec:action in time} 
\begin{figure*}[htb!]
       \centering
       \subfigure[Fuzzy intervals for morning time (blue), having breakfast activity (red), midday (yellow), afternoon (purple), evening (green)]{\includegraphics[scale=0.3]{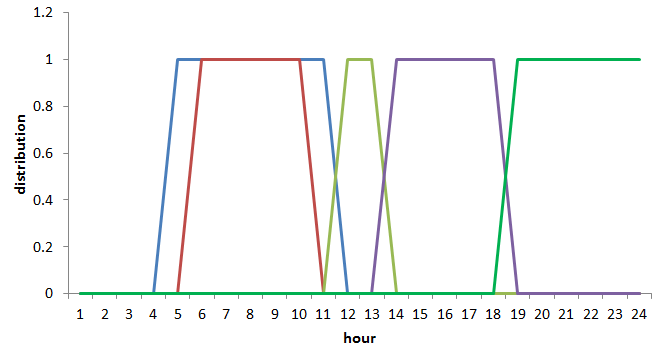}
  \label{gra:fuzzy logics}}
       \quad
       \subfigure[Minimum cartesian distance between the centroid point of the circle with the aggregation radius $r$ and the nearby areas]
       {\includegraphics[scale=0.30]{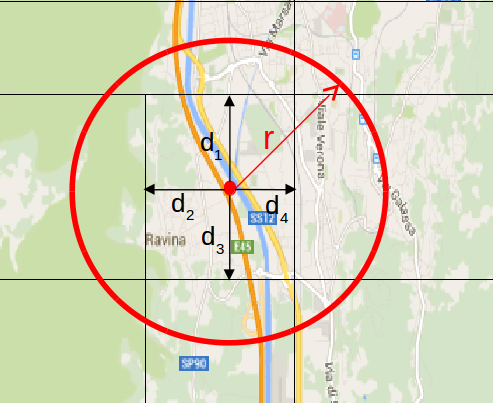}
       \label{gra:radius}}  
       \caption{}
\end{figure*}

The probability of human activity that happens at a certain time and day is $P(a|t,d )$ that is a probability computed using the Fuzzy Reasoning model, as described in the following formula: $ P(a|t,d ) = \frac{S({FM}(a))}{S(FM(t,d))}$, where $S(x)$ is a fuzzy controller which estimates the area of the fuzzy values (fuzzy sets are represented in trapezoidal curve), $FM(a)$ is a trapezoidal fuzzy membership function for possible values of action set \textit{a}; $a \in A$, $A$=\{eating, working, studying,...\} and $FM(t, d)$ is a trapezoidal fuzzy membership function for possible values of timeset \textit{t}; $t \in T$, $T$=\{early morning, mid morning, late morning, mid day,..,late night\}. For example, Figure \ref{gra:fuzzy logics} illustrates that the fuzzy sets for the morning time (interval [6:00 am.,11:00 am.], for the having breakfast activity (interval [6:00 am.,10:00 am.]), for the afternoon time and so on. In this example, the probability of having breakfast in the morning is the intersection between the fuzzy set of morning time period and the fuzzy set of having breakfast. This intersection implies the probability of the action occurring during a certain time period and it is computed as the trapezoid area estimation in the fuzzy controller: $S(x) = \sum\limits_{i=0}^{22}{\frac{x_{i+1} + x_{i}}{2}*h}$, where $h$ is the height of the trapezoid and equal to 1, $a_i$ are the fuzzy set values of action \textit{a}; $a \in A$, $A$ = \{eating, working, studying,...\}, $t_i$ are the fuzzy set values of time period \textit{t}; $t \in T$, $T$= \{early morning, mid morning, late morning, mid day,..,late night\}, $i$ is a parameter for time interval by hour ($i$=0,1,2...,22).

\subsubsection{Likelihood of Human Activity from a set of POIs in a Given Location} \label{sec:action in location}

POIs are the most important factors affecting the likelihood of activity occurrence. But not all the POIs has the same weight. Some POIs, such as bus stops, small shops, ATM, etc., which frequently occur in several areas, have a minor discriminative influence on the likelihood estimation than very distinctive POIs, like an airport or a swimming pool. Borrowing the approach commonly used in the Information Retrieval, we estimate the importance of POIs in a given area by the Term Frequency-Inverse Document Frequency (TF-IDF) \cite{Salton:1988:TAA:54259.54260} function. In this case, the Term Frequency (TF) factor measures the frequency of a POI in a given area, while the Inverse Document Frequency (IDF) factor gives an indication of the general discriminative power of a POI. A high IDF factor is associated with POIs that are rare while a low IDF factor is associated with POIs that are very common and thus are of low usefulness for distinguishing among different areas. For these common POIs, the ratio inside the logarithm of the IDF formula approaches 1, bringing the TF-IDF closer to 0. More in details, the weight of the POIs in location \textit{l} is estimated by the formula, $
\text{tf-idf}(f, l) = \frac{N(f, l)}{\mathop{\rm argmax}\limits_w \{N(w, l) : w \in l\}} * \log \frac{|L| }{ |\{l \in L: f \in l\}|} $, where $f$ is a given POI; $f \in F$, $F$=\{building, hospital, supermarket,...\} and $l$ is a given location; $l \in L$, $L$=\{location1, location2, location3,...\}, $N(f, l)$ is the number of occurrence that POI \textit{f} appears in location \textit{l} \ and $\mathop{\rm argmax}\limits_w \{N(w, l) : w \in l\}$ is the maximum occurrence of all the POIs in location \textit{l}, $|L|$ is the number of all locations, $|\{l \in L: f \in l\}|$ is the number of locations, where POI \textit{f} appears. This way, after we assign a weight to each POI in a given area, we retrieve a set of human activities relevant to those POIs in the area by performing reasoning on the HBOnto ontology. A weight of POIs is propagated to the relevant actions, and the total weight of a given action is estimated as follow: $W(a , l) = \sum_{f \in F(a)} \frac{ \text{tf-idf}(f, l) }{ |{A(f)}| }$, where $a$ is the action in location $l$, $f$ is the element of POI types which derive action $a$, $|{A(f)}|$ is the number of elements (actions) in the set $A(f)$, and $F(a)$ is the set of POI types associated to the action $a$. The probability of human activity that happen in location \textit{l} is a probability given by $P(a|l) = \frac{ W(a, l) }{ W(l)}$, where $W(a, l)$ is the weight of human activity \textit{a} that occurs in location \textit{l}; $l \in L$, $L$ = \{location1, location2, location3,...\}, and $W(l)$ is the total weight of all activities that occur in location \textit{l}. 

\subsubsection{Likelihood of human activity from a set of POIs in or nearby  locations}\label{sec:nearby location}
In some cases, the correct activity for a user cannot be predicted properly based on the POIs immediately close to the user. This happens, for instance, if the user is moving and starts a phone call before reaching his intended destination where the activity will take place. To tackle these cases, we propose an extension of the stochastic model that also takes into account POIs that are not located in the user location but can be find considering a larger neighborhood. 


Starting with the centroid of a location, we consider POIs in locations within 100 m, as a default aggregation radius of the point. If the total number of POI’s in those locations is at lower than \textit{h}=50 (as a default), we expand the aggregation radius by 25 m until the number of POIs in the intersecting locations is higher than \textit{h}, or until the aggregation radius reaches to maximum radius of 3.000 m which was chosen consistently aligned with the maximum coverage of, for instance, mobile phone cells. As an example, Figure \ref{gra:radius} shows that four nearby locations fall inside the aggregation radius \textit{r} of the centroid point of a given location. The weight of the activities in such areas is estimated depending on the intersecting weight [0,1]  of the nearby locations fall inside the circle: $l_i \in L(l,r) $. So we extend $P( a|l) $ using aggregation radius \textit{r}, which is the likelihood estimation of the human activities in given location \textit{l} as the formula: $ P( a|l, r) = \frac{\mathop{\rm \sum}\limits_{l_i \in L(l,r)} \{W( a, l_i ) * \lambda_i\}}{ \sum\limits_{l_i \in L(l,r)} \mathop{\rm \sum}\limits_{a_j \in A(l_i)} \{W( a_j, l_i ) * \lambda_i\}} $, where $W(a,l)$ is the weight of action $a$ in location $l$, and $A(l_i)$ is the set of actions in location $l_i$. The $\lambda_i$ is the intersection weight [0,1] that is measured by the minimum cartesian distance between the centroid point of the circle with radius \textit{r} given by the centroid point of location $l$ and the closest point of nearby areas $\lambda_i = 1-\frac{\{d_i : {d_i} <= r\}}{r} $, where $d_i$ is the distance of nearby location $i$, in which the distance is equal or lower than the \textit{r} of the circle. For instance, if a nearby location falls in the circle, the intersection weight is 1, if not, it is lower than 1. Given $\lambda_i$, the $\mathop{\rm } \{W( a, l_i ) * \lambda_i\}$ is the activity weight in the intersecting locations that fall inside the circle. The following example introduces the human activities with a likelihood measure, which  could be performed in the morning of workday for a given set of POIs in a location as shown in Table \ref{tbl:HRBModel inference}.

\begin{table*}[htb!]
  \centering
	\begin{adjustbox}{max width=\textwidth}
  \begin{tabular}{ | c | c | c | c | l | l |}
    \hline
      \textbf{POIs} &  \textbf{Count} & \textbf{Weight} & \textbf{Activities in a given location}  & \textbf{Activities in the morning of workday} & \textbf{Probability} \\ \hline
     \multirow{4}{*}{$\left<tourism,hostel\right>$} & \multirow{4}{*}{1} & \multirow{4}{*}{0.3} & relaxing at home & relaxing at home & 0.1\\ 
       &  &  & having breakfast & having breakfast & 0.1 \\ 
     & &  & having lunch &  &  \\ 
     & & & having dinner &  &   \\ \hline 
     $\left<natural,tree\right>$ & 10 & 1.5 & hiking & hiking & 0.6\\ \hline 
     $\left<highway,bus\_stop\right>$ & 1 & 0.02 & traveling by bus & traveling by bus & 0.01 \\ \hline 
  \end{tabular}
  \end{adjustbox}
  \caption{The HRBModel inference for human actions in the morning of workday}
  \label{tbl:HRBModel inference}
\end{table*}

\subsection{Usage of the HRBModel}\label{sec:application HRBModel}

The HBRModel estimates the probability of an action to happen in a region on the bases of the frequency of POIs. POIs are bias across areas. Because some locations (i.e., the center of the city) have many POIs that could derive many actions and locations (i.e., the remote area) have few POIs that could lead to few actions. So, we cannot divide a region in areas of the same dimension, but we have to take into account the number of POIs included in an area. We consider the areas to be well characterized in terms of POIs. We divide a territory into grid cells(=locations) in order to have uniformly distributed POIs across locations. We consider Openstreetmap (OSM) as the source of POIs, as it is open and free. We first describe how POI information (human activity relevant) can be extracted from OSM for a given location. Then we described how a territory can be organized into locations satisfying certain properties on the basis of available location and POI data. An default and spatial rectangular grid is designed for populating the representative POIs (i.e., related human activities) at each location. The area of a region or territory is divided into unit areas, where each unit size of the grid is 50 $m^2$ and in each of that we populate the POIs. 

We build a density-based POI distribution grid for a territory which have a threshold number of POIs in each cell(=location) of the grid (in a default configuration, the aggregation radius is 0). The default grid is then re-partitioned using Quad-Trees \cite{QuadTree} in a way that each unit area (=location) contains $h$ $\in$ [10,20] number of POI’s. The Quad-trees are commonly used to partition two dimensional space by recursively subdividing it into four quadrants or regions. The regions might be square or rectangular, or might have arbitrary shapes. We define this re-partitioned grid as a density-based POI distribution grid, in which the locations in urban spaces have a smaller size of coverage as the POI density is higher and the locations in remote spaces have a larger size of coverage as the POI density is lower. In our analysis, we discard and exclude the irrelevant POIs, for example, those do not reflect any human activity like benches, chimneys, power towers. and trash bins. 

\section{Evaluation}\label{evaluation}

In this section, we performed three types of evaluation for the applicability of the predictive model, HRBModel, taking various types of spatial and temporal data and their factors into account: 

\begin{itemize}
  \item Evaluation through user feedback
  \item Evaluation through bank card transaction data revealing actual human economic activities
  \item Extended evaluation of the model 
\end{itemize}

\subsection{Evaluation through user feedback} 

\begin{figure}[htb!]
       \centering
       \includegraphics[scale=0.35]{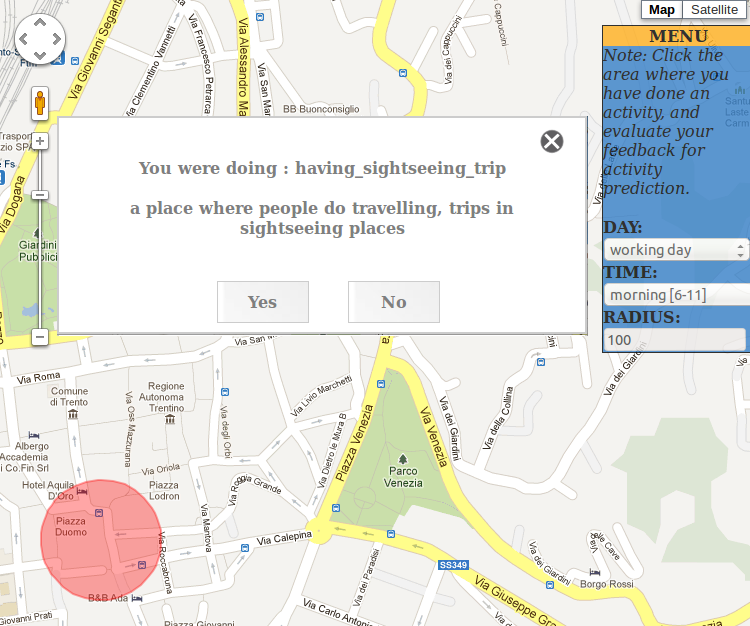}
       \caption{Visualization of demonstration app, at Piazza Duomo on weekday mornings}
       \label{gra:evaluation app}
\end{figure}

We designed an experimental application for collecting user feedback data to validate our HRBModel. The user feedback was collected in the city of Trento, Italy. We then present the experimental outcome and evaluation results. The goal of the experiment was to understand whether human activities associated with a given area at a given time in our fuzzy model are correct, activities are evaluated with respect to the feedback provided by a set of users. The application recommends a list of top-\textit{k} possible human activities at each user-chosen context and then users identified evaluate the correct activity he/she performed from the list. The application shows the map of Trento overlaid by the density-based POI distribution grid of the mobile network coverage map. The interface of the application is illustrated in Figure \ref{gra:evaluation app}. The figure shows the most probable activity in top-\textit{k} is predicted to the user. The preliminary analysis of this evaluation was introduced in our previous work~\cite{DBLP:conf/mum/DashdorjSAL13}: in this work, we improved the fuzzy reasoning model and re-analyzed the geographical area with richer content.  

\begin{figure*}[htb!]
       \centering
       \subfigure[POI distribution in Trento]
       {\includegraphics[scale=0.22]{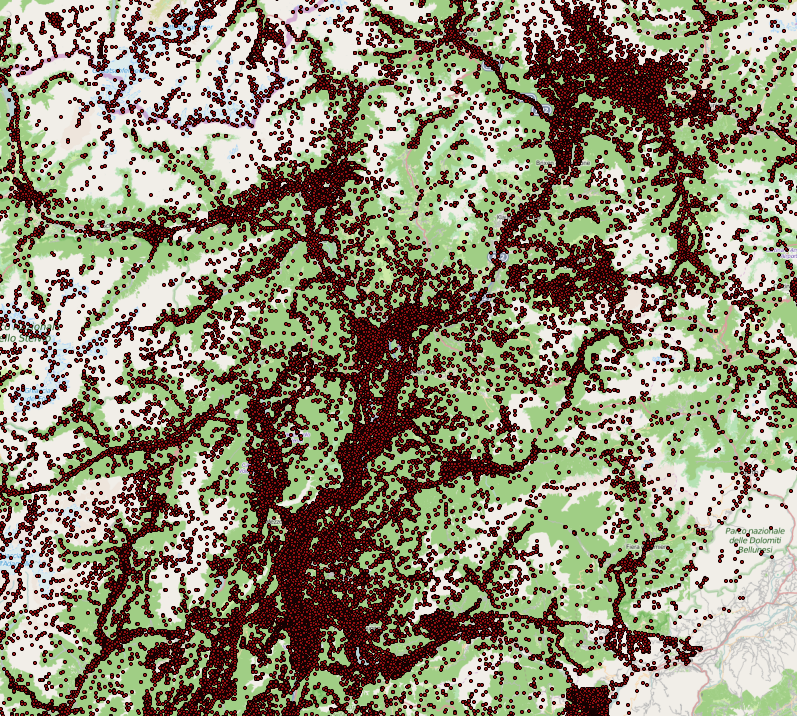}
         \label{gra:POI_dist_Trento}}
       \quad
       \subfigure[Density-based POI distribution grid in Trento]
       {\includegraphics[scale=0.21]{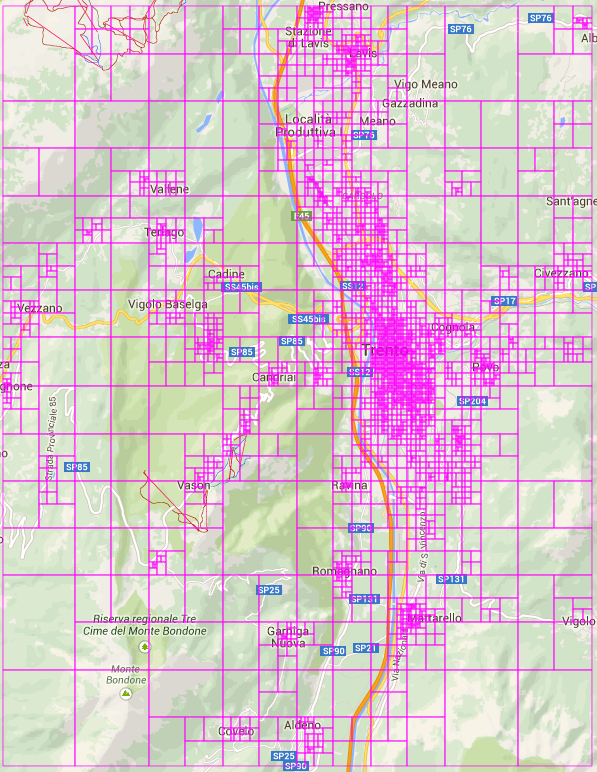}
         \label{gra:Grid_partition}}
       \caption{}
\end{figure*}

In building this application, first of all, we needed to build a density-based POI distribution grid for this city: for this we collected 4.6 million POIs extracted from the OSM. Using Quantium GIS\footnote{www2.qgis.org}, the map of the city has been divided into a grid of dimensions 401 x 302 cells in the beginning, where each unit size of the grid is 50 m$^2$. The cells are partitioned using quad-tree in a way that each unit area (i.e., location) contains \textit{h} $\in$ [10,20] size of representative POIs. The re-partitioned locations containing threshold \textit{h} number of POIs are represented in Figure \ref{gra:Grid_partition}. In total, 3,150 locations were processed. Using the data from OSM, we populated the POIs at each location with those contained in, crossed over located on the border of, or intersected with that location. In total 135,918 relevant POIs were extracted across the locations: this number was reduced to 31,514 after cleaning and discarding irrelevant POIs (i.e., those that do not reflect relevant human activity).

\begin{figure*}[htb!]
       \centering
       \subfigure[The changes of the number of refined POIs (green) and extracted human activities (reddish) compared to the raw POIs, aggregation radius=0]
       {\includegraphics[scale=0.27]{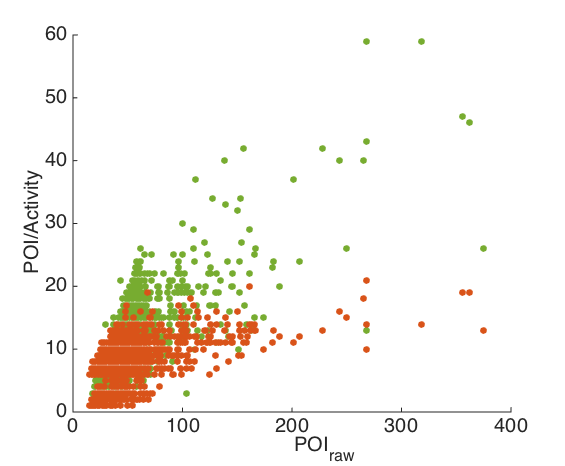}
         \label{gra:scatter_features}}
       \quad
        \subfigure[The distributions of POIs and human activities across locations, aggregation radius=0 (i.e. default configuration)]
       {\includegraphics[scale=0.29]{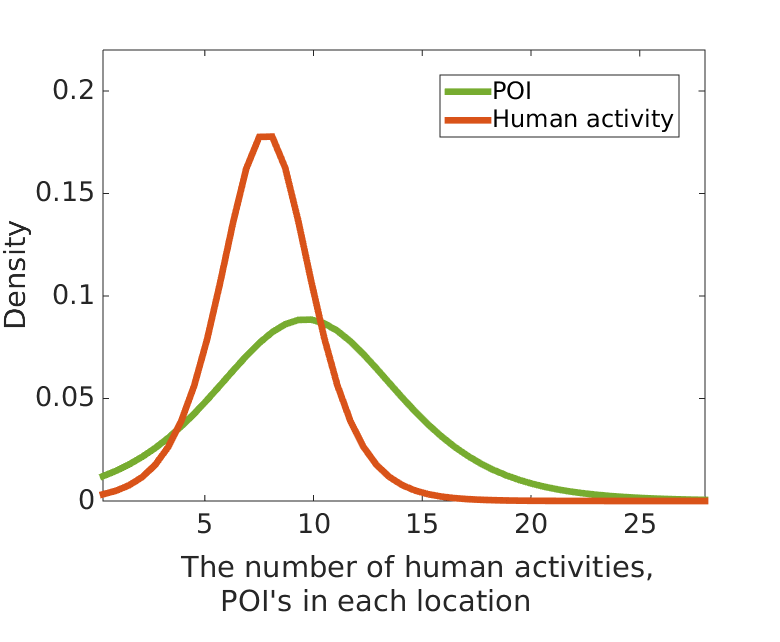}
       \label{gra:POI_dist}}
       \caption{The representative POIs and human activity distribution}
\end{figure*}

Figure \ref{gra:scatter_features} shows the comparison of number changes of the items (i.e. raw and refined POI and extracted human activities) across locations in scatter plot. The refined number of POIs are 10 times lower than the number of raw POIs after the cleaning and discarding irrelevant POIs. The number of human activities extracted from the refined POIs are 5 times lower than the raw POIs after the activity inference. The density distributions of number of POIs and human activities across locations are represented in Figure \ref{gra:POI_dist}, where the density curves are Gaussian types. This figure shows that across locations, the minimum, median, and maximum number of POI are 15, ~38, and 375, and the minimum, median, and maximum number for human activity are 1, ~9, and 59. The remote areas are described in the left tail of the distribution that contain a small number of POIs and human activities, up to the mode of the density distribution of POIs 37, and human activities 9. The dense areas are described in the right tail of the distribution that contain the number of POIs and human activities are up to 375 and 59, respectively. 

\begin{figure} \centering
\begin{tikzpicture}[      
        every node/.style={anchor=north east,inner sep=0pt},
        x=-2mm, y=-2mm,
      ]   
     \node (fig1) at (0,0)
       {\includegraphics[scale=0.38]{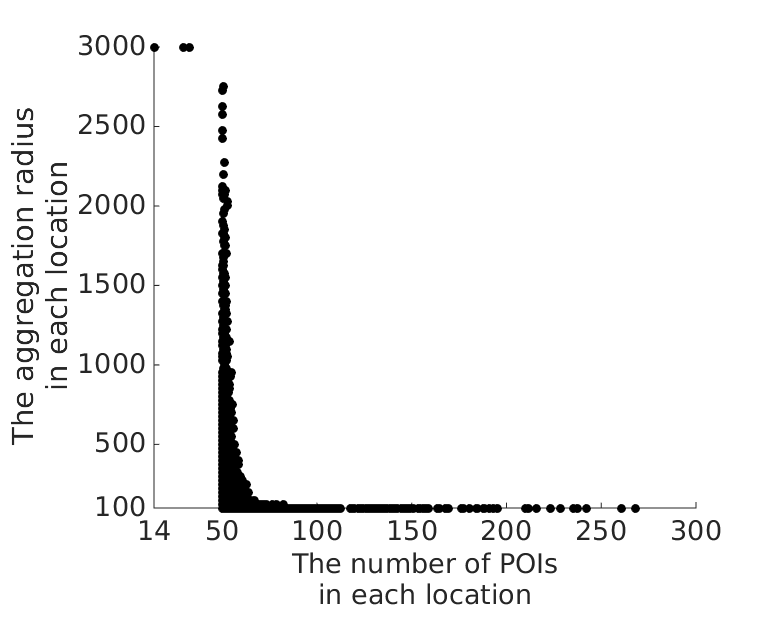}};
     \node (fig2) at (3,3)
       {\includegraphics[scale=0.18]{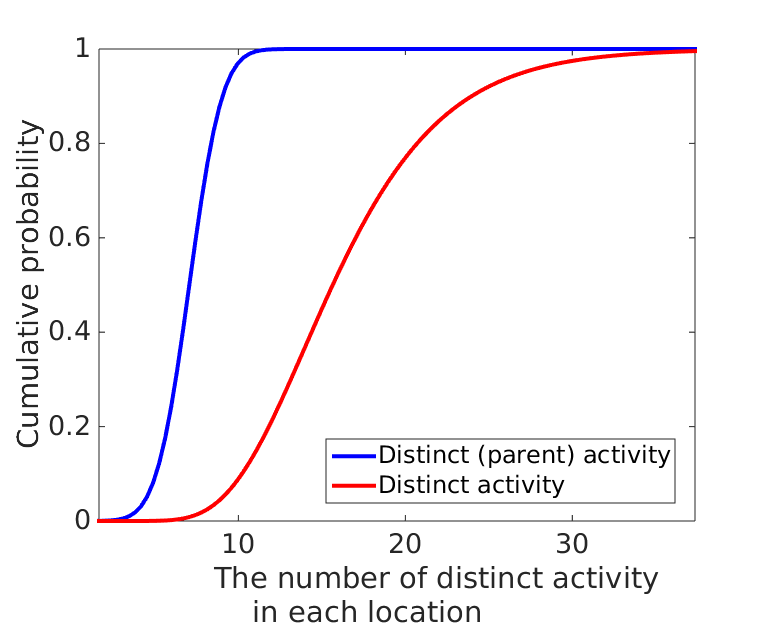}};  
\end{tikzpicture}
\caption{POI and the relevant distinct human activity distribution using aggregation radius}
\label{gra:POI Human activity distribution nearby areas}
\end{figure}

\begin{figure*}[htb!]
       \centering
       \subfigure[Distribution of POI categories across locations of the POI distribution grid, and of locations of the POI distribution grid containing POI categories, aggregation radius=0 ]
       {\includegraphics[scale=0.25]{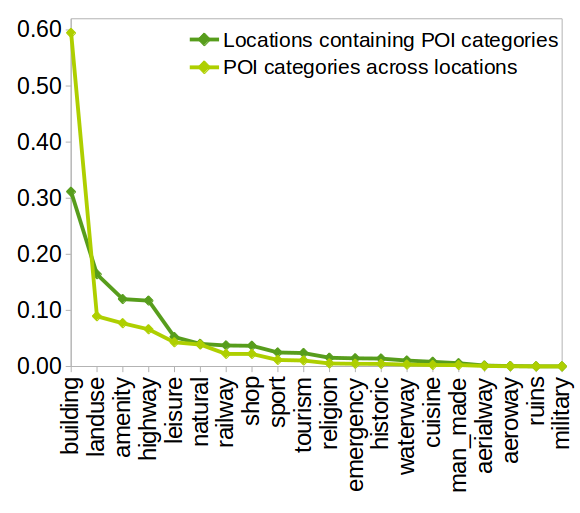}
         \label{gra:act_grid}}
       \quad
        \subfigure[Distribution of activity categories across locations of the POI distribution grid, and of locations of the POI distribution grid containing activity categories, aggregation radius=0 ]
       {\includegraphics[scale=0.25]{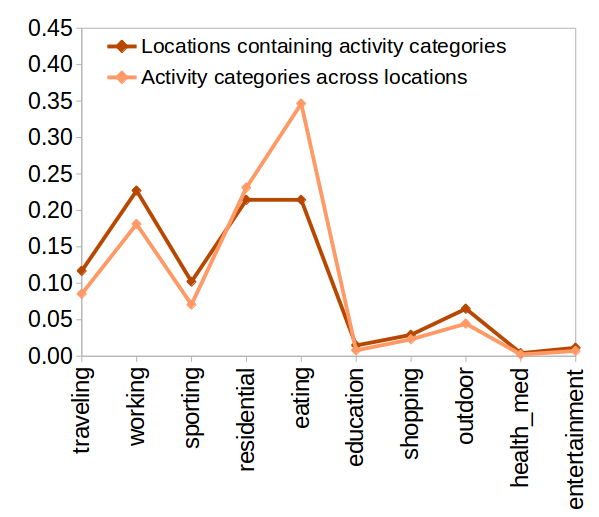}
       \label{gra:poi_grid}}
       \caption{Semantics of POIs and activity distribution over Trento, Italy}
\end{figure*}

To populate a sufficient number and diversity of POIs in each location, we consider the nearby areas for estimating the likelihood of human activities. The nearby areas are the intersected locations within the aggregation radius of the centroid point at each location. The aggregation radius is configured differently for each location: this satisfies the need for the total number of POIs in such intersected locations to be above threshold \textit{h} (see Figure \ref{gra:POI Human activity distribution nearby areas}), where each location has at least \textit{h}(=50) number of POIs in the intersected locations. Across locations, the minimum, median, and maximum number of POIs are 14, 55, and 268. The distinct activities and parent level of distinct activities are estimated in each location using the aggregation radius. The main types of POI in Trento city include building, land-use, amenity, and highway types (see Figure \ref{gra:poi_grid}) and the main human activities extracted from these POIs are mostly related to eating, residential, and work activities (see Figure \ref{gra:act_grid}).   

%

Our web application is able to recommend a set of top-\textit{k} human activities in a given location and time of the user selection. However, multiple activities that have the same priority can occur at the same time. In this experiment, we do not evaluate the order of activities in the prediction due to the limitation of fine-grained ground-truth data we use. So we concentrate only on the ``fuzzification'' of the activity prediction, which evaluates if one of the activities among top-\textit{k} prediction performed by the application is correct or not. However, we can further evaluate the order of the activity prediction if the ground-truth data we use is fine grained. So, a user evaluation among the first \textit{k} predictions can be an initial evaluation to measure the accuracy of the model performance. In any of the above locations, upon selection, the user is prompted with a list of top-\textit{k} human activities associated with the time of day, having the highest likelihood. The user is then asked to select those activities that he or she has actually performed by in that area at that time. The user is also allowed to select an activity not in the top-\textit{k} list among all the available activities, in case none of the above fits. 

We collected user feedback through the web application described above for one week with 32 participants involved. We then analyzed the collected user feedback in order to measure the accuracy of the HRBModel. We measured the accuracy of the model. For the latter, we took into consideration only the areas with the highest amount of feedback, Downtown and Povo. 

To measure the accuracy of the approach, we consider the prediction of an activity correct if the user selected it in the web application regardless of its ordering  position. Within the total of 481 items of users' feedback collected, if we consider a set of the top-8 activities in the recommendation, the overall correctly predicted activity percentage is 68\% (\textit{k}=8). There, we propagated the probability activities to their child activities. We increased the prediction granularity of the activity to \textit{k=n}, where \textit{n} is the total number of distinct activities, and the overall prediction percentage was raised to 93\%. By changing the granularity of activity level to high-level (parent) human activities, the prediction accuracy is increased up to 80\% (\textit{k=8}), and 97\% (\textit{k=n}). When we ignore the time for estimating the likelihood of human activities, the accuracy percentage was lower 57\% (\textit{k=8}) but stable at 97\% for \textit{k=n}. This might be because some human activities are tightly related to the location and have nothing to do with time. For example, if there is an open football stadium, people tend to play there at any time they like. 

So the overall prediction percentage is affected by how we choose the threshold top-\textit{k} of the prediction granularity. This might be a reason that the prediction percentage as well as its granularity can vary at the different local scales. However, the user feedback did not have enough distribution over the city. So we selected two geographical areas; the center of the city and Povo. In these locations, we have sufficient feedback to evaluate the model at the different local scales. The center of the city is characterized with different types of activity, while the Povo is mostly characterized with studying and research work activities as our participants are researchers. We estimated the prediction accuracy in the center of the city, and the result shows that the high-level (parent) activity prediction percentage is 98\% (\textit{k=n}) but 95\% in Povo. This shows that there is an attraction in the city center, so the participants who share a similar work profile or similar study profile go there to perform similar activities. This result matches the finding of Phithakkitnukoon et al~\cite{Phithakkitnukoon:2010:AMI:1881331.1881336}, which showed that a daily pattern of human activity (sequence of human activity) strongly correlates with a certain type of geographic area that shares a common characteristic context.

\subsection{Evaluation through credit card data revealing human economical activities}

We employed user feedback for evaluating the model in the previous section, but the ability to reflect actual human activity is not perfect (note the aforementioned limitations of user feedback data we collected in Trento city). Here, we introduce another source of data that could serve as a direct proxy for human economical activity for a more sophisticated evaluation of our HRBModel. We use bank card transaction data collected in Barcelona, Spain which were performed at POS terminals by users. The type of business activity at every POS terminal, for instance, at a department store or a restaurant, etc., can reveal the human activities that can be performed at such a location. With the use of this data, we evaluate the performance of our HRBModel in a way similar to the evaluation with user feedback in the previous section.

\begin{figure*}[htb!]
       \centering
       \subfigure[Density-based POI distribution grid in Barcelona]
       {\includegraphics[scale=0.25]{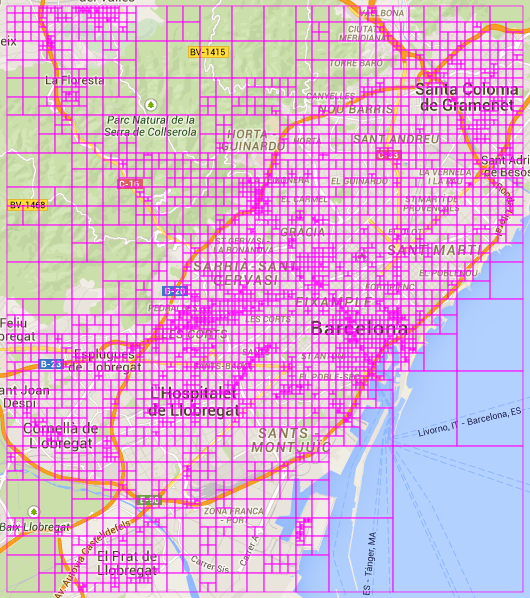}
         \label{gra:Grid_partition_barcelona}}
       \quad
        \subfigure[(Default configuration) POI and human activity distribution across the locations in Barcelona, aggregation radius=0]
       {\includegraphics[scale=0.33]{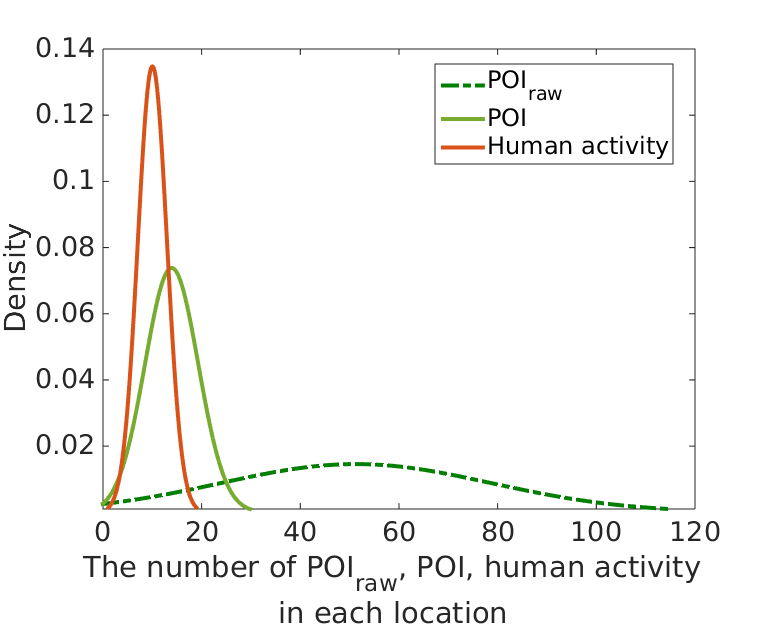}
       \label{gra:POI_dist_barcelona}}
       \caption{}
\end{figure*}

First, we also needed to build a density-based POI distribution grid for the city of Barcelona, Spain and for that we collected 2.6 million POIs extracted from OSM in a rectangular grid with dimensions 252 x 393, where the unit size of each grid is 50 m$^2$. Further, we re-partitioned the grid using the quad-tree approach to ensure that each location contains h $\in$ [10,20] POIs. Final partitioning is presented in Figure \ref{gra:Grid_partition_barcelona}. In total, 3,853 locations were processed. Using the data coming OSM, we populated the POIs that were contained in, crossed over, were located on the border of, or intersected each location. In total, 197,289 activity relevant POIs were extracted across the locations, which was reduced to 53,529 after the cleaning and discarding of irrelevant (POIs those that do not reflect any human activity). Figure \ref{gra:POI_dist_barcelona} summarizes the POI filtering results compared to the raw POIs. In this case the POIs are 5 times lower compared to the number raw POIs in each location. The figure also shows the corresponding human activities to POIs that were generated after the raw POI cleaning process. The POI and human activity density distribution across the locations are represented in Figure \ref{gra:POI_dist_barcelona}, where the density curve is skewed in the center. The figure also shows that across locations, the min, median, and max number of POI are 1, ~47, and 545, and the min, median, and max number for human activity are 1, ~13, and 46. The remote areas contain a small number of POIs and human activities, up to the mode of the density distribution of POIs 32 and human activities 12. In dense areas, the number of POIs and human activities are up to 545, and 46, respectively. These POIs and human activities are the representative activities in each location. 

\begin{figure}
\centering
\begin{tikzpicture}[      
        every node/.style={anchor=north east,inner sep=0pt},
        x=-2mm, y=-2mm,
      ]   
     \node (fig1) at (0,0)
       {\includegraphics[scale=0.38]{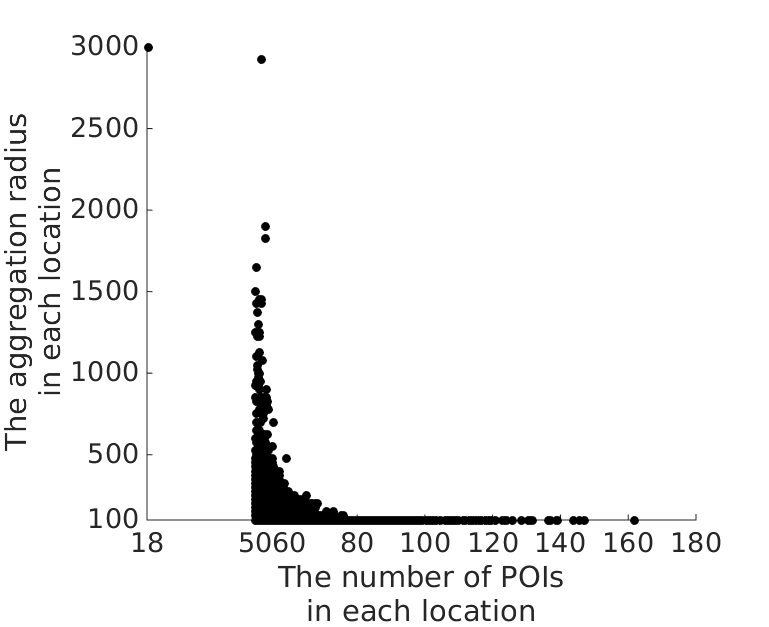}};
     \node (fig2) at (3,3)
       {\includegraphics[scale=0.18]{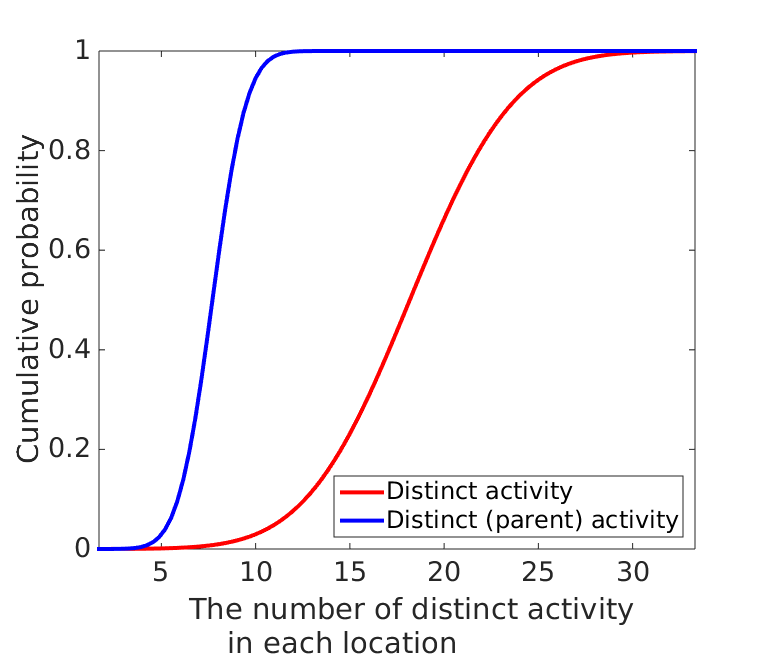}};  

\end{tikzpicture}
\caption{POI and the relevant distinct human activity distribution considering aggregation radius}
\label{gra:Barcelona POI Human activity distribution nearby areas}
\end{figure}

To populate a sufficient number and diversity of POIs in each location, we consider the nearby areas for estimating the likelihood of human activities. The nearby areas
are the intersected locations within the aggregation radius of the centroid point at each location. The aggregation radius is configured differently in each location, which satisfies the need for the total number of POIs in such intersected locations to be above the threshold \textit{h} (see Figure \ref{gra:Barcelona POI Human activity distribution nearby areas}) where each location has at least \textit{h}=50 number of POI’s in the intersected
locations. Across locations, the minimum, median, and maximum number of POIs are 18, 55, and 162. The distinct activities and parent level of distinct activities are estimated in
each location using the aggregation radius.

%

\begin{figure*}[htb!]
       \centering
       \subfigure[Distribution of POI categories across locations of the POI distribution grid, and of locations of the POI distribution grid containing POI categories, aggregation radius=0 ]
       {\includegraphics[scale=0.25]{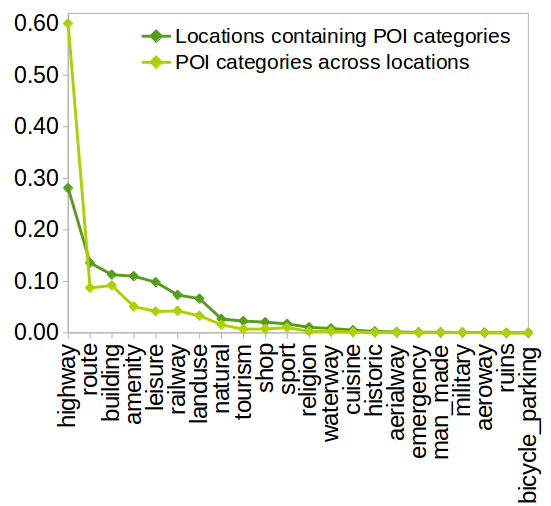}
         \label{gra:act_grid_Barcelona}}
       \quad
        \subfigure[Distribution of activity categories across locations of the POI distribution grid, and of locations of the POI distribution grid containing activity categories, aggregation radius=0 ]
       {\includegraphics[scale=0.25]{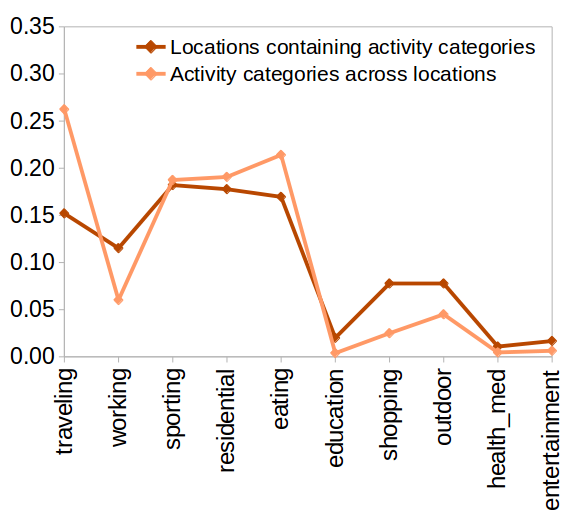}
       \label{gra:poi_grid_Barcelona}}
       \caption{Semantics of POIs and activity distribution over Barcelona}
\end{figure*}

The main types of POI in Barcelona city are highway, building, and route types (see Figure \ref{gra:poi_grid_Barcelona}) and the main human activities extracted from these POIs are eating, residential, and sporting (see Figure \ref{gra:act_grid_Barcelona}).

We performed the evaluation of the model to measure the accuracy of the approach in same way as we have done in Trento city, but here we selected random POS locations in Barcelona; assume that each POS generates correct human activities. Every POS is associated with a business category: therefore, 76 types of businesses are given for all the POSs, such as travel, food, hypermarkets, hotels, real estate, automation, bars and restaurants, personal care, sport and toys, technology, home, content, fashion, leisure, health, transport, etc. Since we use both POIs and POSs in the data sets in our evaluation, we adjust them so that they are comparable to avoid a semantic gap between the data sets. For instance, when a customer buys groceries in a supermarket, POS interprets that a customer is doing shopping in grocery stores. The POIs and POSs use different activity type taxonomies, so they need to be mapped to a common taxonomy for comparability. We define consistent activity types for POS using the human activities described in ActOnto. We discard POIs and POSs that have no activity type reflected in it, such as ATM use. 

For evaluating the HRBModel, we compared the human activities of a given POS with the set of top-\textit{k} human activities generated by HRBModel in the same context of POS location and time, using the aggregation radius of nearby areas for estimating the likelihood of human activities, as we described in Section \ref{methodology}. We performed this validation by selecting 1,000 random POSs in the city. The overall accuracy of parent level activity prediction percentage is 82\% (\textit{k}=8) and 92\% (\textit{k=n}). When we ignore the time, the parent level activity prediction percentage is stable at 92\% (\textit{k=n}). This also shows that some human activities are tightly related to the location and have nothing to do with time. However, the random selection of POS is not a sophisticated choice for analyzing the granularity of the prediction task in the city scale. We need to estimate the accuracy fraction at the local scales to know where the HRBModel gives better accuracy and good precision. 


%

\subsection{Extended evaluation of the model}

\begin{figure}[htb!]
  \centering
  \subfigure[POI distribution in Barcelona]{\includegraphics[scale=0.15]{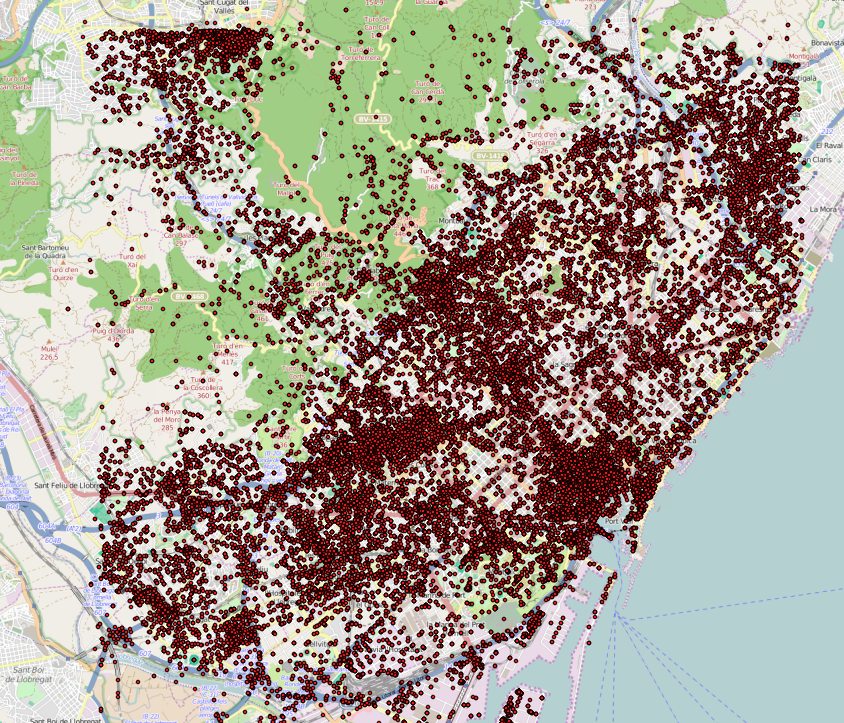}
  \label{gra:regular}}
  \quad
  \subfigure[POS distribution in Barcelona]{\includegraphics[scale=0.15]{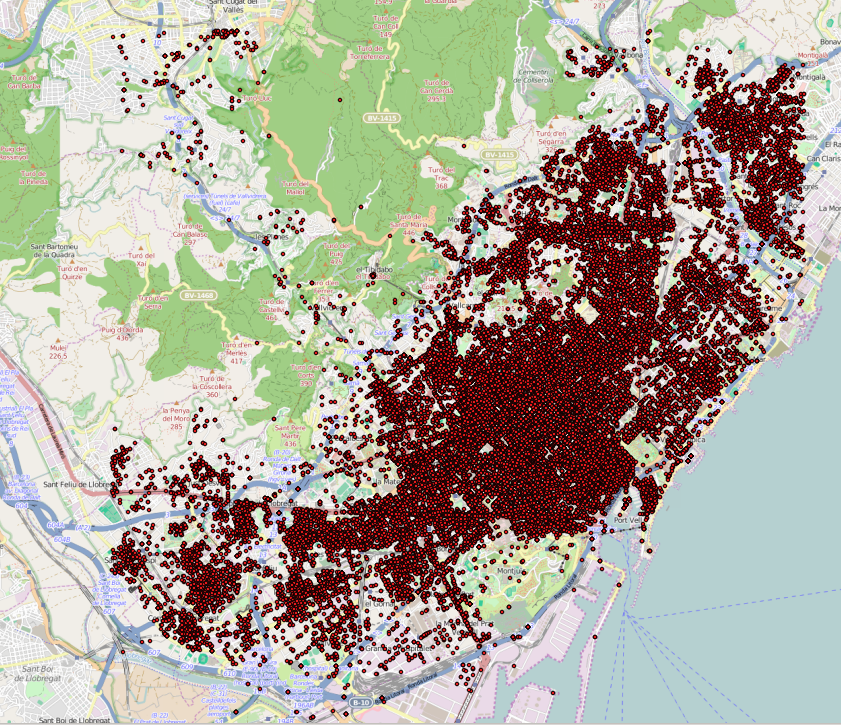}
  \label{gra:irregular}}
  \caption{POI and POS distribution in Barcelona}
  \label{gra:POI POS distribution Barcelona}
\end{figure}

In this section, we consider an extensive validation of the HRBModel in order to estimate if POI data can be a good proxy for predicting such a specialized type of activity (economical activity) at the local scale. For that purpose, we compare the possible activity distribution predicted by our model for each location across the city with the actual data on the observed spending activity in POS at that location. This gives us ground-truth information on for instance, what types of economic activity are occurring in a city. However, spatial data (i.e., POI and POS data) distribution (see Figure \ref{gra:POI POS distribution Barcelona}) is very uncertain and heterogeneous, and extracting meaningful information from it is a very challenging task. The data can be imprecise and not available in some areas. The heterogeneity of the data directly influences the confidence regarding how the data is characterized. We performed city-wide normalization of POI activity distribution to match that of POS, and then matched the different categories of human activities in the POI and POS data, which gives good accuracy and good precision both at the local scale and for different land-use types.

In our analysis, POS data reflects spending activity on a fine-grained scale, economic activities, which are very important component of spending activity. The POI data, on the other hand, is coarse grained, and each of its points represent a single coarse-grained geographical object, not the objects within it. For example, POS machines are associated with every store at the mall, and the type of POS represents a type of activity performed at the store. But in POI data, the mall could be represented as a single building of type “mall”, which does not contain any information about stores within the mall. This granularity difference causes a quantitative difference in the categories of activities in between the POI and POS data. These differences could introduce a substantial bias in the model performance. To achieve comparable performance between the data sets, which would enable using POI data for predicting economic activity, we first need to remove the overall systematic bias of the data sets. We introduce a global normalization to make POIs and POSs consistent at the macro-level in order to account for possible heterogeneity in the representativity across different categories of activities in the two data sets. 

\begin{figure}
\centering
\begin{tikzpicture}[      
        every node/.style={anchor=north east,inner sep=0pt},
        x=-2mm, y=-2mm,
      ]   
     \node (fig1) at (0,0)
       {\includegraphics[scale=0.38]{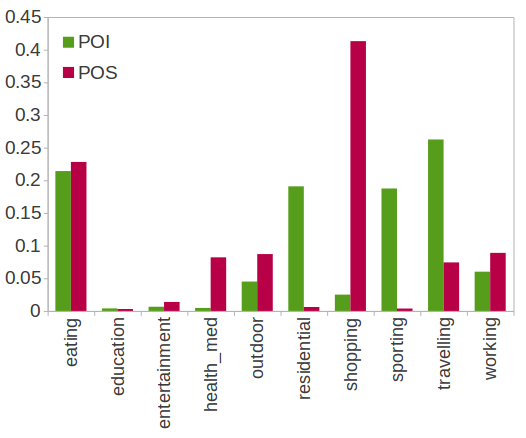}};
\end{tikzpicture}
\caption{Overall economical activity distribution in each data set, 1) POI 2) POS}
\label{gra:activity_distribution_POS_POI_new}
\end{figure}

We generate the distribution of different human activities in POI and POS data as depicted in Figure \ref{gra:activity_distribution_POS_POI_new}, where non-economic activities including educational, outdoor, residential, sporting, and working types of activity, have less of a value in the both distributions of different activities in POI and POS data. An economic activity is a special but very important component of spending and here we hypothesize that POIs can be used to predict the economic activity after appropriate normalization (i.e., $W_{poiNorm}$) for the general data set representativity bias on the city scale. The result might also be associated with the fact that spending is just one of the components of human activity and not all activities have something to do with spending.

\begin{figure*}[htb!]
\centering
  \subfigure[Variation of aggregation radius in each location]{
  \resizebox{6cm}{!}{
  	\begin{tikzpicture}[ 
        every node/.style={anchor=north east,inner sep=0pt},
        x=-2mm, y=-2mm,
      ]   
     \node (fig1) at (0,0)
      {\includegraphics[scale=0.28]{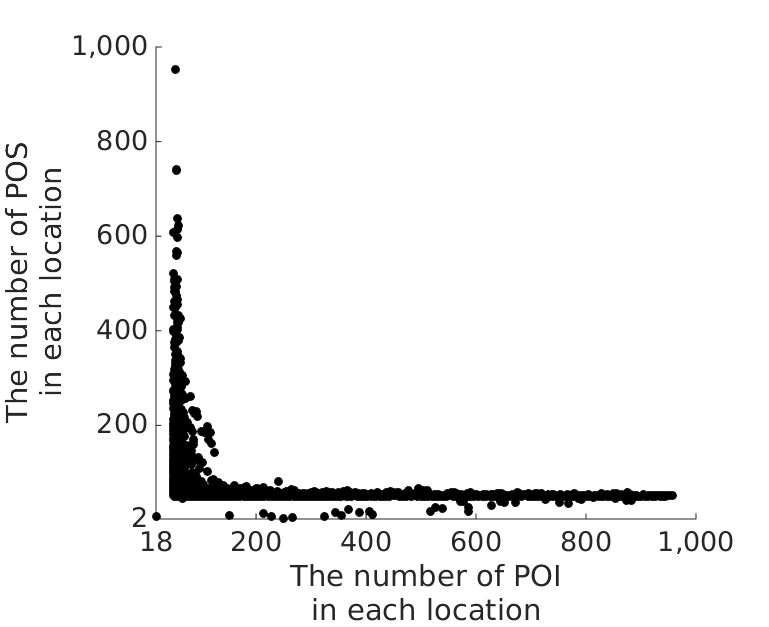}\label{gra:aggregated POI POS}};
     \node (fig2) at (3,3)
       {\includegraphics[scale=0.13]{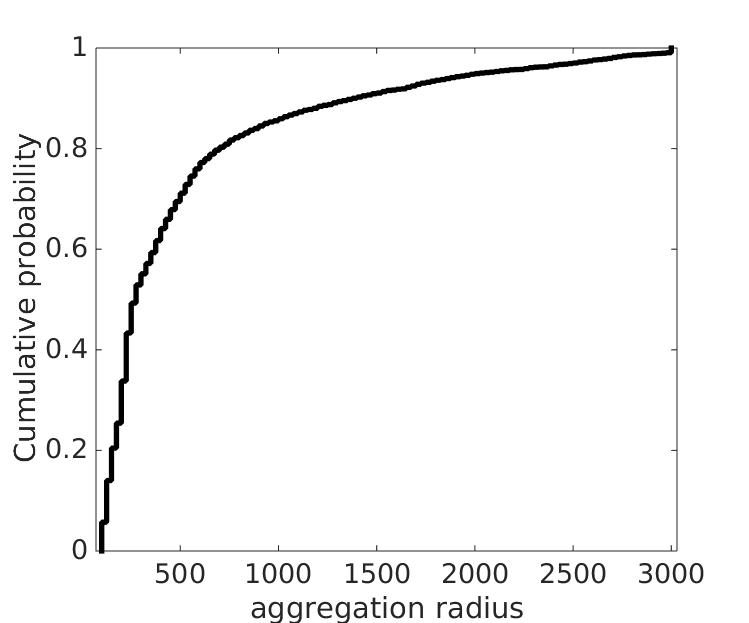}};  
\end{tikzpicture}
}
}
       \quad
  \subfigure[Variation of the location length after the aggregation]
       {\includegraphics[scale=0.27]{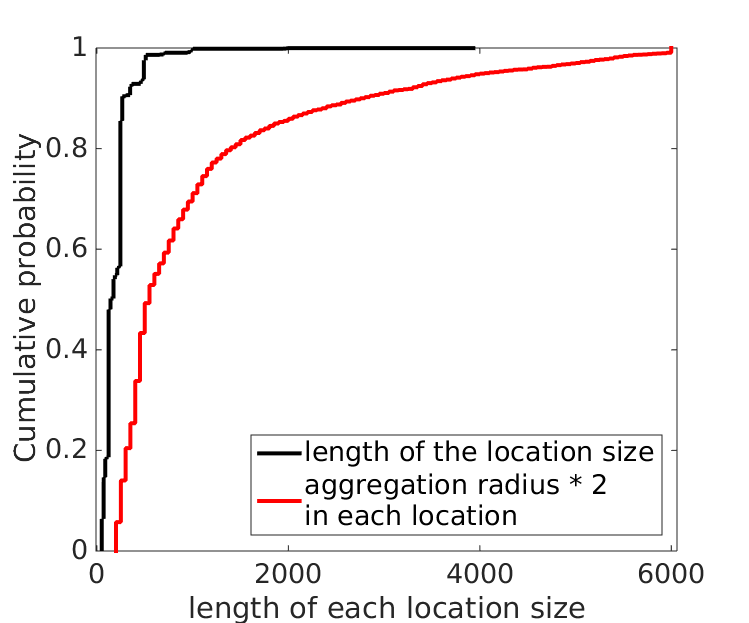}\label{gra:aggregated location length}}
  \caption{}
\end{figure*}

\begin{figure*}[htb!]
       \centering
  \subfigure[POI and POS distribution across the location in the aggregation radius]
       {\includegraphics[scale=0.28]{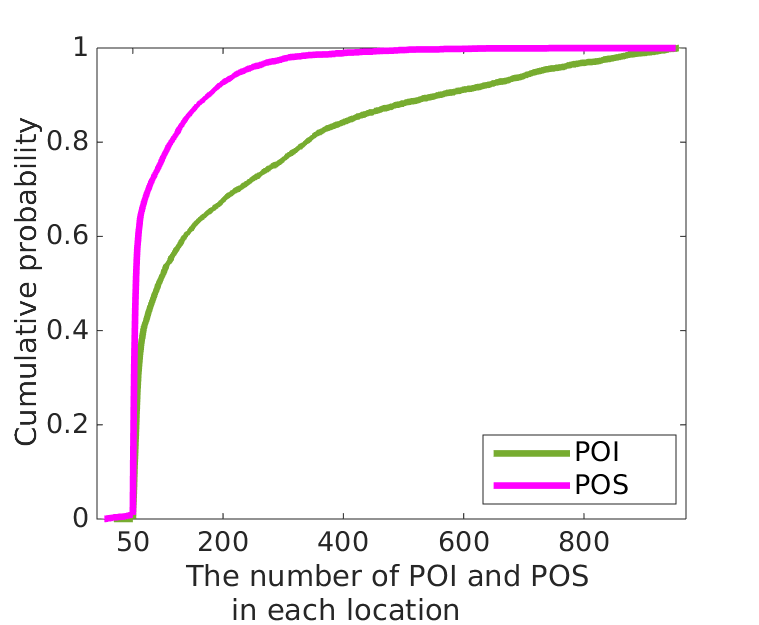}\label{gra:aggregated POI POS distribution}}
       \quad
  \subfigure[The POI and POS normalized estimate error vs local estimate error]
       {\includegraphics[scale=0.28]{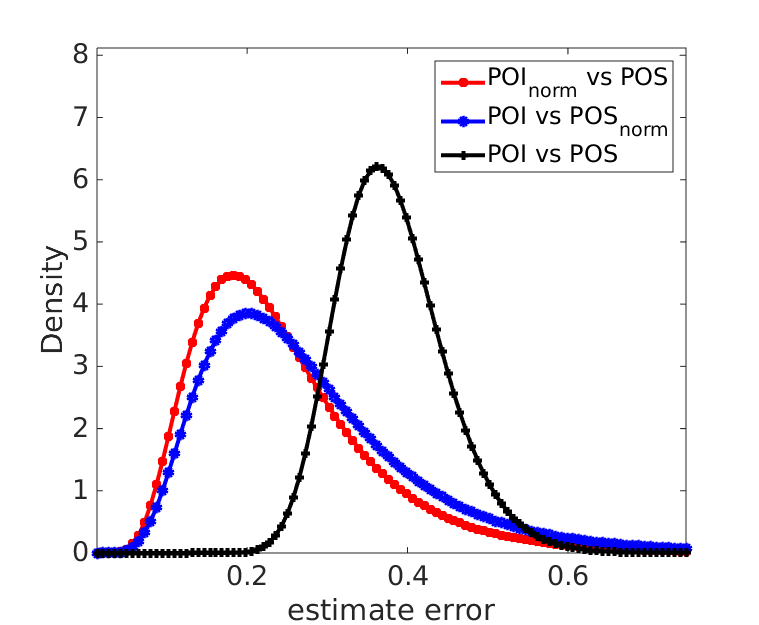}\label{gra:error_bias}
       }
	  \caption{}
\end{figure*}

For each location \textit{l} we have certain distribution of POIs among \textit{n} categories of activity we consider with relative weights, $W_{poi}(a, l)$ and estimate the likelihood of human activities extracted from POIs starting with the centroid of location, considering the POIs in all locations within 100 m of the point, as we described in Section \ref{methodology}. Here we use a larger radius size in case there are not enough POIs within a considered circle: we expanded the aggregation radius by 25 m if fewer than 50 POI’s and POSs were contained in intersecting locations, or until the radius reaches to 3.000 m. This also gives a sufficient number of POIs in remote areas. The aggregation radius selected in each location is presented in Figure \ref{gra:aggregated POI POS}. The length of each location size is processed in cumulative probability after the aggregation radius selection, as plotted in Figure \ref{gra:aggregated location length}. So, 80\% of total locations are increased to the length of 2,000 m. Figure \ref{gra:aggregated POI POS distribution} shows the number of POIs and POSs at each location using the cumulative probability function. Across locations, the minimum, median, and maximum number of POI are 18, ~93, and 958, and are 3, ~55, and 952, respectively, for POS.

Now before comparing the values of $W_{poi}$ with $W_{pos}$, we look at the overall distributions first, (i.e., $W_{poi}(a, c)$ vs $W_{pos}(a, c)$ where $c$ is the entire city). It might happen that those average relative weights are also not the same if there exists a systematic bias introduced by the way POS and POI are defined (for example, it could happen that each restaurant appears in POIs but only 50\% of restaurants appear in POS (say, not all of them accept cards) and this situation could change from one category to another. The overall POI and POS distribution across the city is depicted in Figure \ref{gra:activity_distribution_POS_POI_new}, where the global distributions are relatively similar for the type of activity labeled, eating. So, if this systematic bias exists, then it is no surprise that we will also see a difference for any particular location, which would reduce our reported accuracy.

In order to account for this bias we consider the following normalization: 
\\$W_{poiNorm}(a,l)=\frac{W_{poi}(a,l)*W_{pos}(a,c)}{W_{poi}(a,c)}$ and for each location we compare $W_{pos}(a,l)$ vs $W_{poiNorm}(a,l)$ (which is already normalized by the possible systematic bias between POS and POI distributions) instead of $W_{poi}(a,l)$. In each location, the bias between $W_{pos}(a,l)$ vs $W_{poiNorm}(a,l)$ is compared by the Hellinger distance \cite{Sengstock:2011:ECG:2093973.2094017,nikulin2002hellinger,Sengstock:2013:PMS:2525314.2525353} measure, which quantifies the similarity between two probability distributions and lies between 0 and 1. So we call it as an estimate error between the two distribution: \\
$HD(W_{poiNorm}, W_{pos}) = \frac{1}{\sqrt{2}} \; \sqrt{\sum_{i=1}^{k} (\sqrt{W_{poiNorm}(i,l)} - \sqrt{W_{pos}(i,l)})^2}$, where \textit{i} is an activity of \textit{n} category in each location.

The estimate error between the distribution of activities in POI and POS data in a given location is called \textit{local estimate error}, and the estimate error between the distribution of activities in $POI_{norm}$ and $POS$ in a given location is called \textit{POI normalized estimate error}. The estimate error between the distribution of activities in $POI$ and $POS_{norm}$ in a given location is called \textit{POS normalized estimate error}. The density distribution of local estimate errors, and the POI and POS normalized estimate errors are described by logistic distribution fit across all the locations within the city as depicted in Figure \ref{gra:error_bias}. The peak (mode) of the POI normalized estimate error gives the smallest error values compared to the local and POS normalized estimate error values. The peak (mode) of the POI normalized estimate error distribution density falls at 0.18, which represents an 82\% match in accuracy between the POI and POS data. Either way, POS normalized estimate error would give the same distribution, but the mode of the distribution is lower. On the other hand, the peak (mode) of the local estimate error distribution density falls at 0.40, which is around 60\% accuracy on the POI and POS match. So if we make two distributions comparable by de-biasing them globally, the global estimate error is approximately reduced by 22\%, which is the quantitative difference between the POI and POS data sets. For the general data set representativity bias on the city scale (which might also be associated with the fact that spending is just one of the components of human activity and that not all activities involve spending). From the results of previous section, we note that POI data can be used as a proxy to predict human activities after appropriate normalization. We will show that once overall city normalization is performed, two data sets start to give comparable estimates for human activity at the local scale: this means that one needs to know only the overall bias on the city scale and then POI data can be applied to predict actual human activities. As POS reveals actual human economic activity, we propose to use POI to estimate economic activity when POS data is not available.

\begin{figure*}[htb!]
  \centering
  \subfigure[The number of POIs in each random location of different land-use types]
  {\includegraphics[scale=0.27]{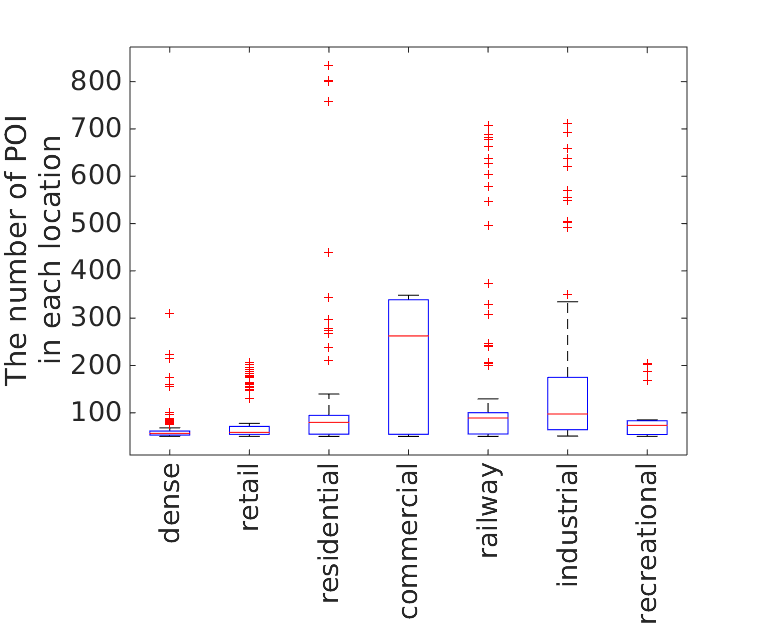}
   \label{gra:error range landuse POI}}
   \quad
  \subfigure[The number of POSs in each random location of different land-use types]
  {\includegraphics[scale=0.27]{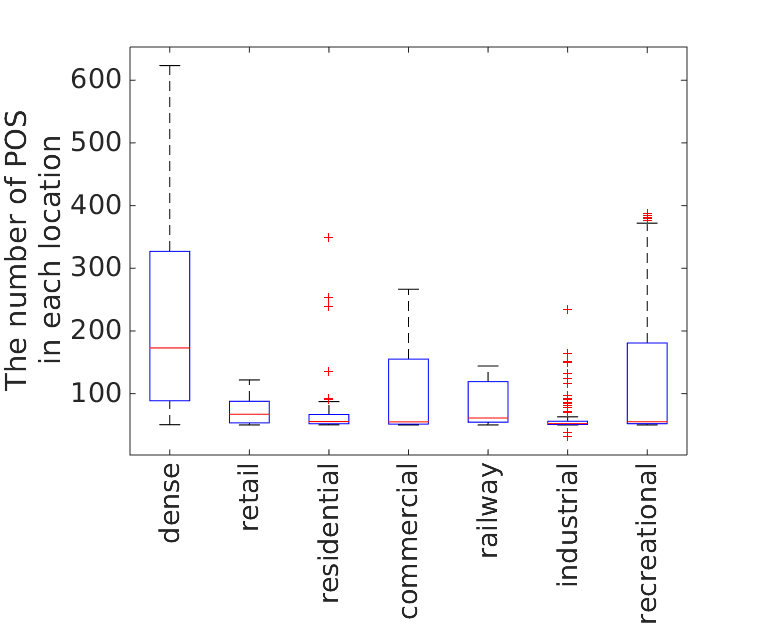}
   \label{gra:error range landuse POS}}
  \caption{The number of POI and POS in each random location selected in different land-use types}
\end{figure*}

\begin{figure*}[htb!]
  \centering
  \subfigure[Variation of aggregation radius selected of random locations in different land use types]{
  \resizebox{6cm}{!}{
  	\begin{tikzpicture}[      
        every node/.style={anchor=south east,inner sep=0pt},
        x=-2mm, y=2mm,
      ]   
     \node (fig1) at (0,0)
      {\includegraphics[scale=0.29]{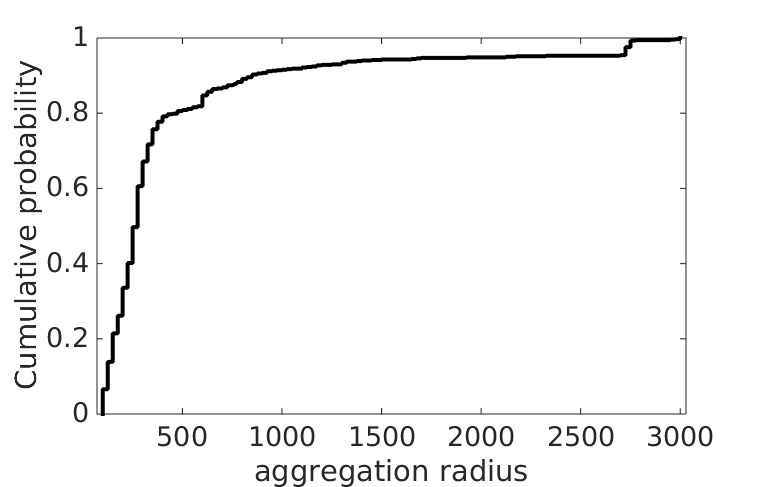}\label{gra:error range landuse aggregation radius}};
     \node (fig2) at (3,3)
       {\includegraphics[scale=0.14]{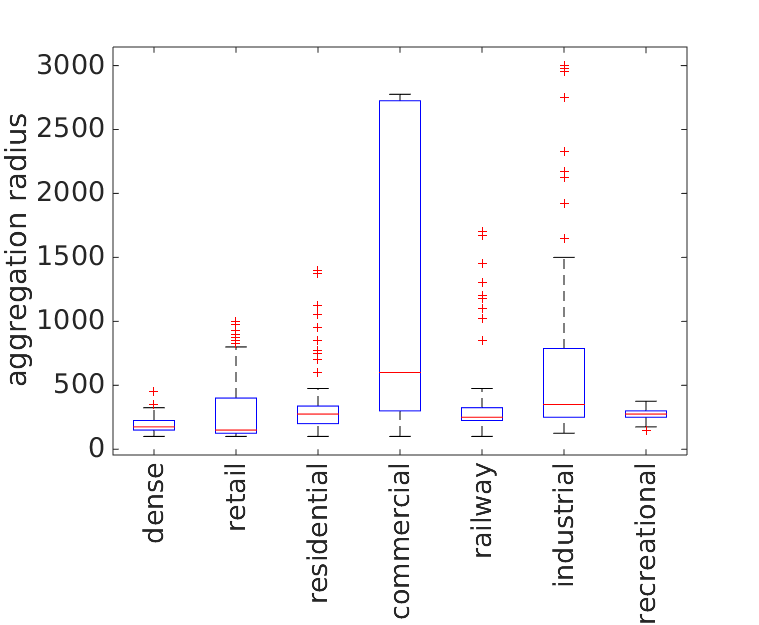}};  
    \end{tikzpicture}
    }
    }
 \subfigure[Variation of an overall estimated error of economical activities prediction with respect to POI normalized and actual POS activities] 
  	{\includegraphics[scale=0.27]{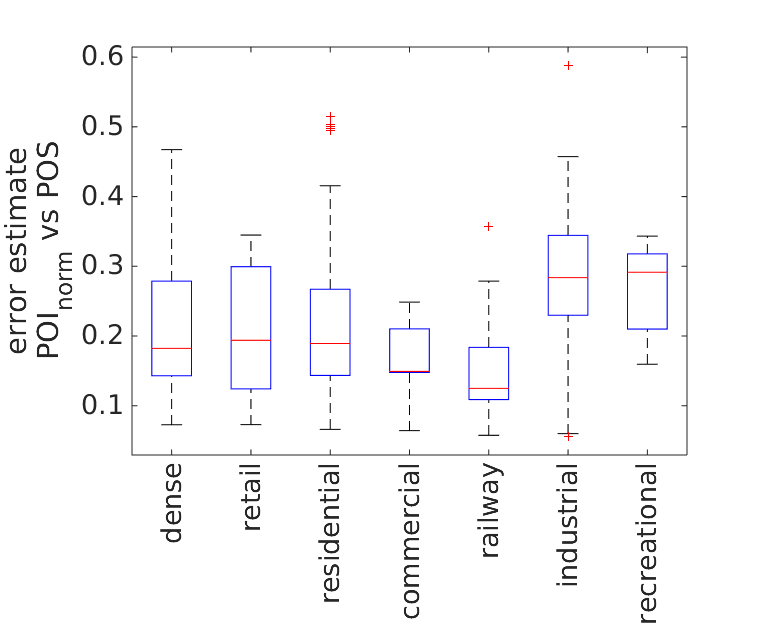}\label{gra:error range landuse}}
  	\caption{}
\end{figure*}


\begin{figure*}[htb!]
  \centering
  \subfigure[Differences in predicted levels of different types of activity across various locations vs observed activity levels in dense (the center of the city) land-use based on POS data]
  {\includegraphics[scale=0.27]{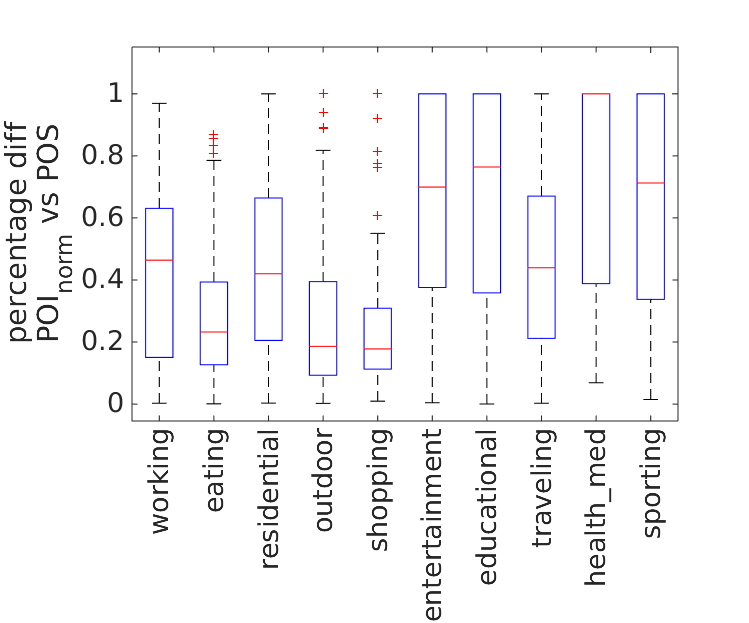}
   \label{gra:error_bias_activity dense}}
   \quad
  \subfigure[Differences in predicted levels of different types of activity across various locations vs observed activity levels in railway land-use based on POS data]
  {\includegraphics[scale=0.27]{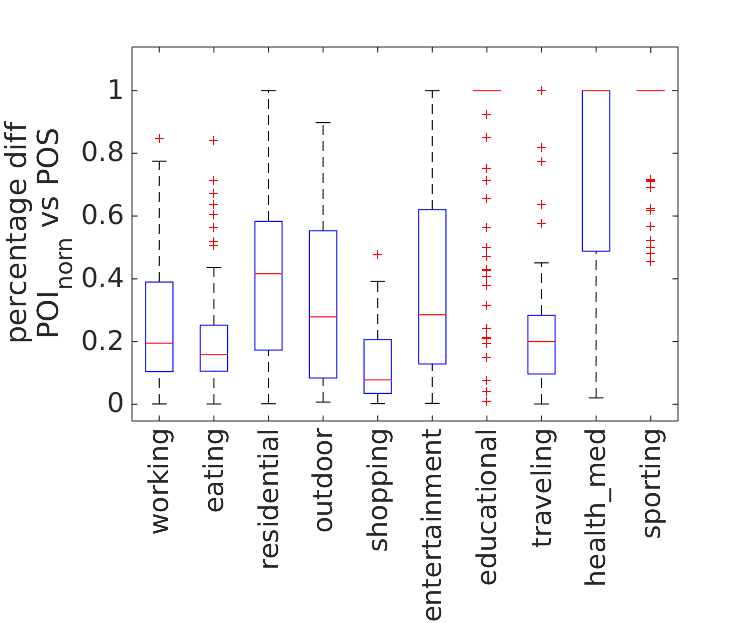}
   \label{gra:error_bias_activity railway}}
  \caption{}
\end{figure*}

As we have seen above, the estimate error distribution is of Gaussian type and on the left tail, the prediction percentage is better (around 84-95\%), while the right tail is heavy tailed down to 20\%. This could be an effect of land-use types, where the distribution of activities in POI and POS data is well described in some type of land areas and not in others. For instance, while POS data about economic-type activity, it is supposed to have strong match in railway, commercial and dense areas. Therefore, we investigate the POI normalized estimate error distribution in different land-use types, which are referenced by OSM: industrial, recreational, commercial, railway, retail, residential, and the central part of the city. We randomly selected up to 100 locations in each land-use type, which were chosen in order to be consistent with the previous work in \cite{Dashdorj:2014:HAR:2675316.2675321}. The populated sufficient POIs and POSs in each location (see Figures \ref{gra:error range landuse POI}, \ref{gra:error range landuse POS}) using the aggregation radius (see Figure \ref{gra:error range landuse aggregation radius}). We excluded the outlier locations whose aggregation radius was more than 1,000 m, which removed 9\% of the total locations randomly selected. 

The POI normalized estimate error distribution in those land-use types is shown in Figure \ref{gra:error range landuse}, where the railway, commercial, and dense areas show lower error values with the accuracy of (81--90\%): this is to be expected, since we validated the POI data with the POS data, which is mostly related to shops. In the railway area, we got the lowest error values, with an accuracy of 88--90\%. In a dense area (the center of the city), the POI normalized estimate error with the accuracy of (81-84\%), and the error distribution is significantly heavy tailed as the area consists of a wide range of human activities. The commercial land-use is also well described with the accuracy of 86\%, while the industrial and recreational ground types of land-use have lower accuracy than 72\%. In order to estimate the sensitive error values among the different categories of activities at those local distributions of POI and POS data, for instance, in dense and railway areas, we estimated the POI normalized percentage difference, which is an error between $W_{poiNorm}(a,l)$ vs $W_{pos}(a,l) $ among the different categories of activities. The percentage difference is a difference between two values divided by the average of the two values, then $PD(W_{poiNorm}(a,l), W_{pos}(a,l)) = \frac{|W_{poiNorm}(a,l)-W_{pos}(a,l)|}{W_{poiNorm}(a,l)+W_{pos}(a,l)}$. Figures \ref{gra:error_bias_activity dense} and \ref{gra:error_bias_activity railway} show the POI normalized percentage difference among the different categories of activities in dense and railway areas. In dense areas, among the categories, the shopping activity has the lowest error (around 11--17\%) in the distribution of activities between the POI normalized and POS as shops are mostly located in the center of the city. Also in railway areas, the shopping activity has the lowest error (around 3--7\%) in the distribution of activities as we evaluated economic-type activities from POI and POS. The health and medical activity has an error value of almost 100\%, as actions of this type are not usually described in POS data. However, while the analysis suffers from a lack of POIs and POSs information collected the model applicability error can be more meaningful for estimating the data loss. The representativity of the data sets by the global normalization at the city scale allows us to estimate how well POI data can predict human activity at the global scale, and the result was validated in land-use types (e.g., railway and dense land-uses).

\section{Possible developments (extensions) }\label{usage of the model}

We presented in previous sections that the HRBModel was developed in order to infer human activities by extracting concepts and their relations from open geographical data. In this section, we briefly explain how the correlation between context and human behavior encoded in HRBModel, can be exploited to infer human behavior from CDR data. To this purpose, we enrich CDR with human activities, which are derived from our model through location and time.

\begin{figure}[htb!]
  \centering
  \subfigure[regular coverage]{\includegraphics[scale=0.12]{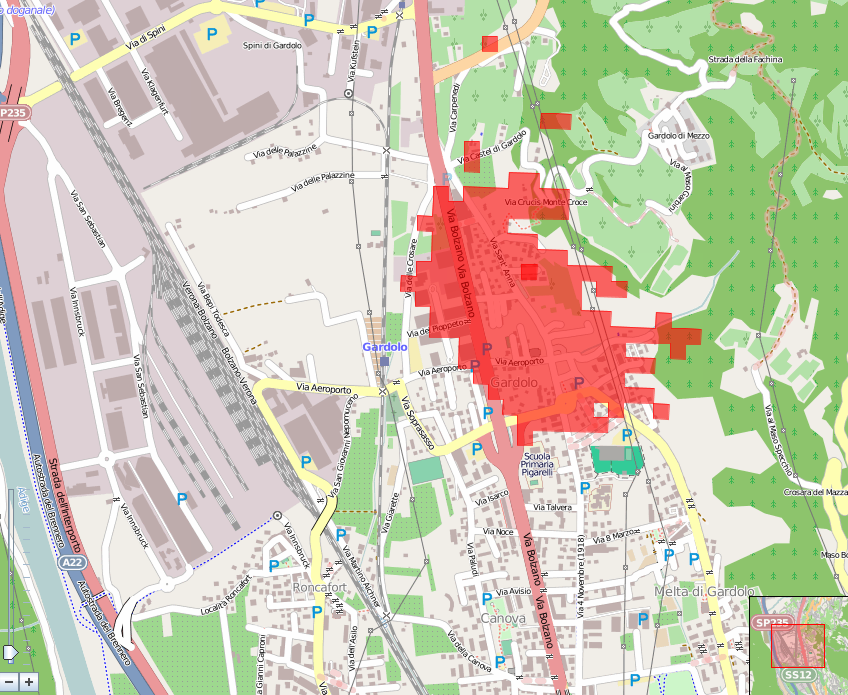}
  \label{gra:regular}}
  \quad
  \subfigure[irregular coverage]{\includegraphics[scale=0.12]{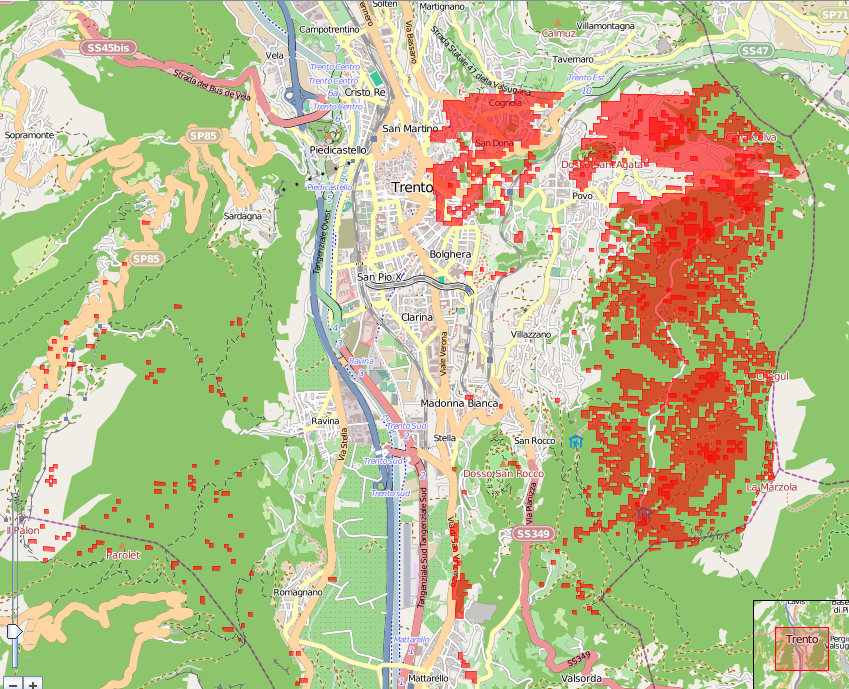}
  \label{gra:irregular}}
  \caption{Example of the two extreme cases of cell coverage area}
  \label{gra:coverage-areas trento}
\end{figure}

\begin{figure*}
       \centering
       \includegraphics[scale=0.25]{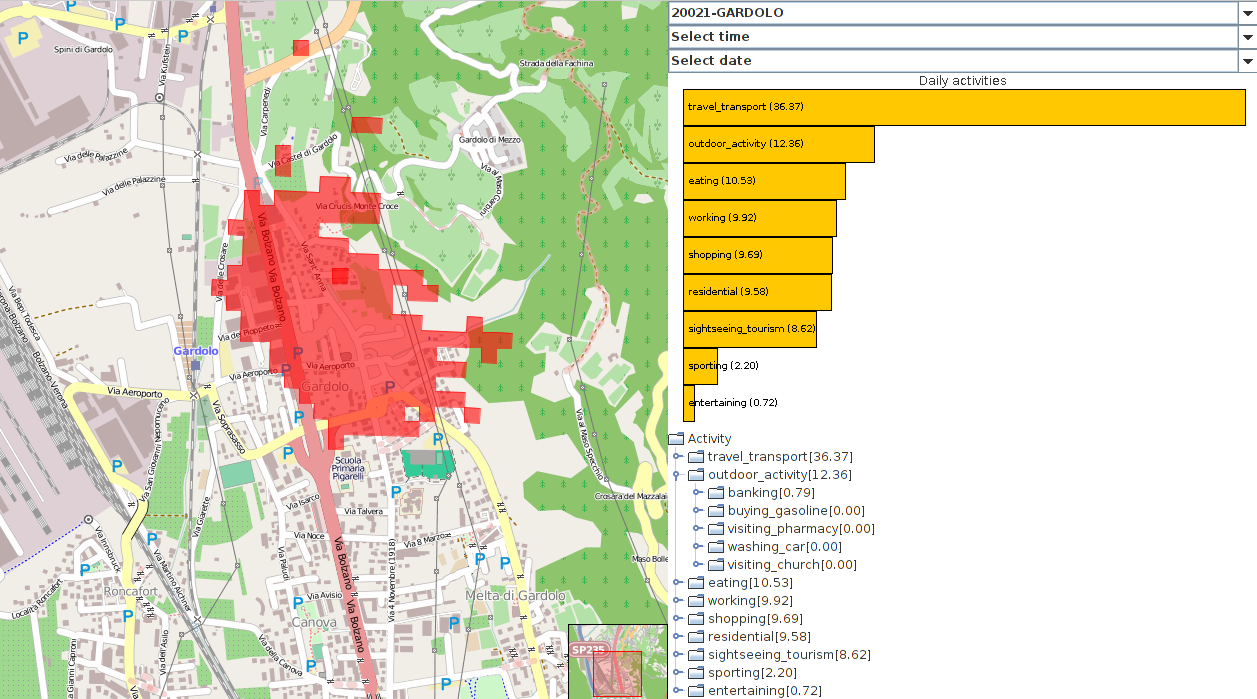}
       \caption{Visualization of the ranked activities in the cell coverage for Gardolo area of Trento }
       \label{gra:experimental app}
\end{figure*}

As we mentioned earlier, our main contribution of this article is to use background knowledge regarding human activities that could happen in the context where mobile network events occur \cite{Zolzaya:DC:2013,Zolzaya:Poster:2013}. The most common mobile network antennas are GSM 900 and GSM 1800. Each antenna covers a certain area for the support of mobile communication. Sometimes, cell coverage areas are not precisely defined, and they can be temporarily modified depending on the estimation of call traffic from/to an area. Usually, the size of the coverage areas is inversely proportional to the density of the population inhabiting the area. However, there is an upper bound of the coverage area size due to antennas physical limits to 35 km. Concerning the shape of the cell coverage areas one can observe that,
when the territory is regular (i.e., flat with no mountains or other natural irregularities), the shape of cells can be approximated with convex polygons, but in presence of irregular
territory with a lot of mountains and valleys, the region associated
to a cell can be very irregular and possibly disconnected. Irregular coverage areas makes particularly difficult the task of
estimating the calling point. In other cases, the shape of cells can be very irregular and possibly disconnected. An example of these two extreme cases are shown in Figure \ref{gra:coverage-areas trento}. Irregular coverage areas make the task of estimating the calling location particularly difficult. Regular coverage is a contiguous area, and irregular coverage forms a less distinct area with many holes and irregularities. The HRBModel can provide qualitative description of human activities associated with a likelihood measure, to each cell coverage area. For example, a cell coverage area of Gardolo in Trento city is characterized with the possible human activities ranked that could happen in that area, associated with its likelihood, which is illustrated in Figure \ref{gra:experimental app}. In this case, travel by transport, outdoor activity, and eating are higher ranked activities.

Contextually enriched CDR can be exploited to analyze and evaluate call patterns\footnote{A call frequency pattern is a quantitative model that describes the behavior of a communication in a certain class of contexts} (i.e., human mobility, communication and interaction patterns) associated to human activities for revealing existing relationships between human behaviors, and CDR stream. This can be done through a combination of the HRBModel and the bottom-up approaches in pattern extraction tasks used in the state of the art. The qualitative and quantitative properties in semantically enriched CDR allow us to generate training data with a certain accuracy that, together with other behavioral data coming from bank card transaction, social networks and so on, can be used as ground-truth information for training those supervised learning approaches for automatic activity inference in the area of mobile phone data analysis.

Like many factors that influence human behavior, the CDR related to such behavior varies depending on the different situations. For example, these situations could involve a behavioral pattern of attending a football match on rainy day or sunny day, or doing so in the central area or in remote area. Another example of the contextual factors that influences human behaviors is a social event. For instance, human behavior is different during national or local holidays. While some people used to work during national holidays, others may have vacation plans to go somewhere. This concept has been successfully applied in \cite{D4D:2013} where we created an event repository that collected data from the Ivory Coast city in order to analyze CDRs for the identification and characterization of human behavior patterns in the city. Event is an another key element that can be used to enrich the background knowledge of mobile phone data records. 

Therefore, the HRBModel is very useful for filling the missing data in spatial data-sources, where such information is unavailable and measuring the noise, scalability and accuracy among the spatial data-sources. For example, if economical type of activities are not sufficient enough when generated in bank card transaction data, we could combine the OSM data with bank card transaction data in order to better study economical or social and emerging types of activity ~\cite{RePEc:sae:urbstu:v:49:y:2012:i:7:p:1471-1488, D4D:2013,DBLP:journals/corr/SobolevskySGCHAR14,sobolevsky2014money} at the macro level. The most directly applicable scenario includes, for instance, at the finest scale, a semantic enrichment of cell coverage areas, trajectories, and  behavioral patterns (normal or exceptional), and at the broad scale, semantic interpretation of CDR traffic data more toward smart (intelligent) city research that gain better understanding of real-life cases, resulting in better classification of human behaviors in a region or territory. 

The HRBModel can be enabled for a further consideration of the GeoSparql extensions\footnote{http://geosparql.org/} in order to represent and querying of geospatial linked data, such as geo-objects and geo-actions for the Semantic Web. The geospatial RDF/OWL data which can support both quantitative spatial reasoning on the Semantic Web and querying with the SPARQL database query language. Further, we enrich the HRBModel with other contextual information like weather information, events and some other types of linked objects and data. This possibility to extend the model further helps us to refine human behavior recognition by considering more complex and complete contextual information (via linked open data) by employing real-time inference techniques some of which could be learned and decoupled with the pre-processing of features from the raw data of mobile phones.

\section{Conclusion}\label{conclusion} 
Even without taking a person's environment into account, many factors influence the collective human behavior: the geographical information system is certainly only one of these factors. We proposed a model able to enrich mobile phone data records with the context of human activity. The qualitative terms of the human activities can be used for labeling the data in order to train a statistical model for automatic human behavior recognition tasks given heterogeneous and uncertain CDR data. By leveraging data from Web 2.0 as a source of contextual information for mobile data records, we study the possibility to infer human activities with a certain accuracy. Our model is a combination of ontological and stochastic model, HRBModel, for predicting a set of human activities associated with a likelihood measure in a given context of mobile phone data records. Our study demonstrated that semantic expressiveness of human-environmental relationships could be investigated, in principal, based on diverse geographical information systems. The evaluation of the model was performed in two different cities and it demonstrates the applicability and scalability of the model under the conditions of heterogeneity and uncertainty of those spatial data sources like POI and POS data. Furthermore, this model takes various types of spatial and temporal factors into account. Based on our evaluations performed in two different cities, the prediction level depends to some extent on the granularity of the activity being predicted. The analysis shows that the level of activity prediction from the POI data is significant (around 84--95\%). As we concentrated on the model evaluation for a specific economic type of activities using point of sale bank card spending data, we validated that within areas of different land-use types, the POI is a good proxy for predicting those economic activities with 81--90\% accuracy in dense, railway, and commercial areas. This way the context of mobile network records can be used to deepen the context aware human activity recognition from the mobile network events. Hence, a number of challenges remain for future research and deeper investigations of human and environmental dynamics, such as behavior recognition and classification from the semantically enriched CDR. 

\section{Acknowledgements}\label{acknowledgements} 
The authors would like to thank the Telecom Italia Semantic Knowledge Innovation Lab (SKIL) and Banco Bilbao Vizcaya Argentaria (BBVA) for providing the datasets for this research. The authors further thank the Fondazione Bruno Kessler Data Knowledge Management Unit (FBK-DKM) as well as Ericsson, the MIT SMART Program, the Center for Complex Engineering Systems (CCES) at KACST and MIT CCES program, the National Science Foundation, the MIT Portugal Program, the AT\&T Foundation, Audi Volkswagen, BBVA, The Coca Cola Company, Expo 2015, Ferrovial, Liberty Mutual, The Regional Municipality of Wood Buffalo, UBER and all the members of the MIT SENSEable City Lab Consortium for supporting the research. 

\bibliographystyle{spbasic}      
\bibliographystyle{spmpsci}      
\bibliographystyle{spphys}       
\bibliography{ref}   


\end{document}